\RequirePackage{fix-cm}
\documentclass[twocolumn]{svjour3}          

\smartqed  

\usepackage{times}
\usepackage{epsfig}
\usepackage{graphicx}
\usepackage{amsmath}
\usepackage{amssymb}
\usepackage{colortbl}
\usepackage{color}
\usepackage{xcolor}
\usepackage{csquotes}
\usepackage{multirow}
\usepackage{multicol}
\usepackage[export]{adjustbox}
\usepackage{xspace}
\usepackage{soul}
\usepackage{gensymb}
\usepackage{csquotes}
\usepackage{hyperref}
\usepackage{amsmath}
\usepackage{nccmath}

\usepackage{graphicx}
\newcommand{\latinphrase}[1]{\textit{#1}} 
\newcommand{\etal}{\latinphrase{et~al.}\xspace}
\newcommand{\ie}{\latinphrase{i.e.}\xspace}

\newcommand{\eg}{\latinphrase{e.g.}\xspace}
\newcommand{\etc}{\latinphrase{etc.}\xspace}

\newcommand{\ch}{\checkmark}
\begin{document}

\title{Insert your title here
}

\title{Attention Based Real Image Restoration}

\author{Saeed~Anwar \and Nick~Barnes \and Lars~Petersson}


\institute{S. Anwar \at
is a research scientist with Data61-Commonwealth Scientific Industrial Research Organization (CSIRO) and lecturer at School of Computer Science and Engineering, Australian National University, Australia.\\
\email{saeed.anwar@data61.csiro.au}
\and
N. Barnes \at 
is an Associate Professor with the Research School of Engineering and Computer Science, Australian National University, Australia. 
\and
L. Petersson \at
is a principal research scientist and group leader at Data61-Commonwealth Scientific Industrial Research Organization (CSIRO), Australia.
}

\date{Received: date / Accepted: date}

\maketitle

\begin{abstract}
Deep convolutional neural networks perform better on images containing spatially invariant degradations, also known as synthetic degradations; however, their performance is limited on real-degraded photographs and requires multiple-stage network modeling. To advance the practicability of restoration algorithms, this paper proposes a novel single-stage blind real image restoration network (R$^2$Net) by employing a modular architecture. We use a residual on the residual structure to ease low-frequency information flow and apply feature attention to exploit the channel dependencies. Furthermore, the evaluation in terms of quantitative metrics and visual quality for four restoration tasks \ie Denoising, Super-resolution, Raindrop Removal, and JPEG Compression on  11 real degraded datasets against more than 30 state-of-the-art algorithms demonstrate the superiority of our R$^2$Net. We also present the comparison on three synthetically generated degraded datasets for denoising to showcase our method's capability on synthetics denoising. The codes, trained models, and results are available on \url{https://github.com/saeed-anwar/R2Net}.
\end{abstract}

\section{Introduction}
\label{sec:introduction}
In the present digital age, many hand-held camera-based devices allow humans, machines, \etc, to record or capture video and image data.  However, during image and video acquisition, various forms of corruption, for example,  noise (Gaussian, speckle, thermal \etc), compression (JPEG \etc), blur (motion, defocus \etc) are often inevitable and can downgrade the visual quality considerably. The process of reducing the artifacts to recover the missing details is called image restoration. Moreover, being a low-level vision task, image restoration is a crucial step for various computer vision and image analysis applications such as computational photography, surveillance, robotic vision, recognition, and classification, \etc

Generally, restoration algorithms can be categorized as model-based and learning-based. Model-based algorithms include non-local self-similarity (NSS)~\cite{Dabov2007BM3D,Buades2005NLM}, sparsity~\cite{Gu2014WNN,peng2012rasl}, gradient methods~\cite{xu2007Iterative,weiss2007makes}, Markov random field models~\cite{roth2009fields}, and external restoration priors~\cite{anwar2017category,Yue2014CID,luo2015adaptive}. The model-based algorithms above are computationally expensive, time-consuming, unable to suppress spatially variant degradations directly as well as characterize complex image textures. On the other hand, discriminative learning aims to model the image prior from a set of degraded and ground-truth image sets. One technique is to learn the prior steps in the context of truncated inference~\cite{chen2017TNRD} while another approach is to employ brute force learning, for example, CNN methods~\cite{zhang2017DnCNN,zhang2017IrCNN}. CNN models improved restoration performance thanks to their modeling capacity, network training, and design. However, the performance of current learning models is limited and tailored for specific synthetic degradation types. 
 
A practical restoration algorithm should be efficient, flexible, perform restoration using a single model and handle both spatially variant and invariant degradations when the degradation is known or unknown. Unfortunately, the current state-of-the-art algorithms are far from achieving all of these aims. We present a CNN model that is efficient and capable of handling synthetic and real degradation present in images. We summarize the contributions of this work in the following paragraphs.
 
\begin{figure}
\begin{center}
\begin{tabular}[b]{c@{ }c@{ }c} 
      
\includegraphics[trim={5cm 2cm  5cm  2cm },clip,width=.15\textwidth,valign=t]{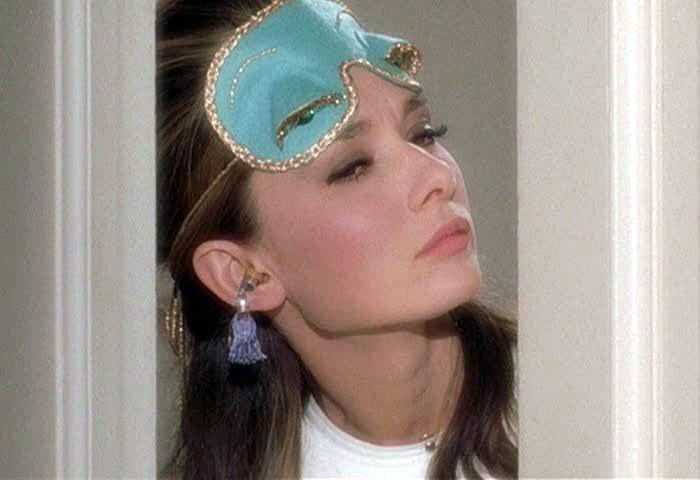}&   
\includegraphics[trim={5cm 2cm  5cm  2cm },clip,width=.15\textwidth,valign=t]{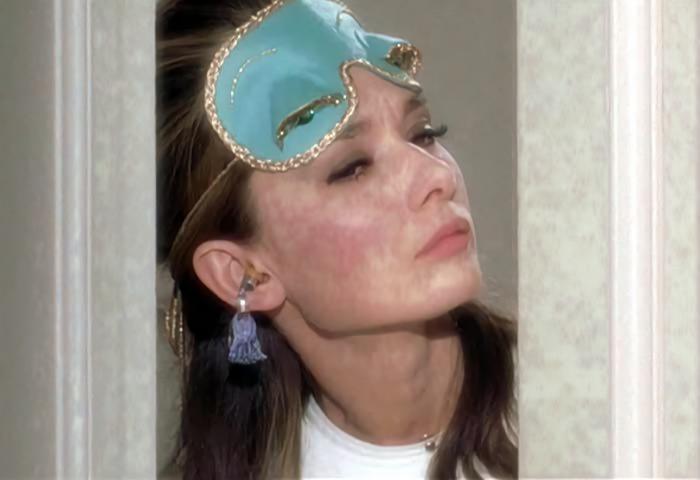}&
\includegraphics[trim={5cm 2cm  5cm  2cm },clip,width=.15\textwidth,valign=t]{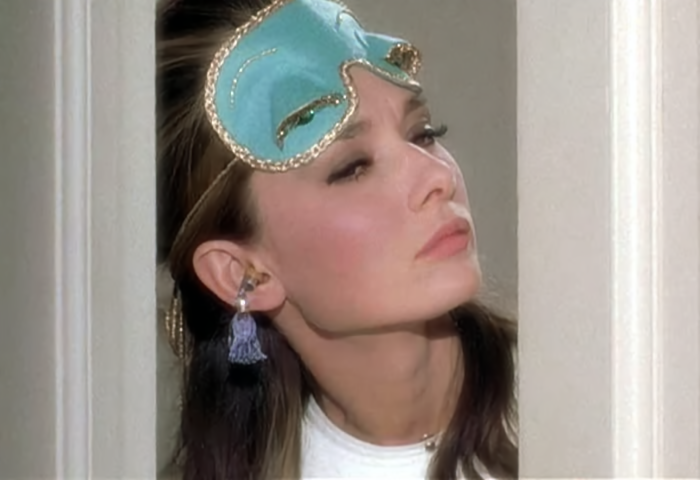}\\
Noisy & CBDNet~\cite{guo2018CBDnet} & R$^2$Net (Ours)\\
\end{tabular}
\end{center}
\caption{A real noisy face image from the RNI15 dataset~\cite{lebrun2015NC}. Unlike 
CBDNet~\cite{guo2018CBDnet}, R$^2$Net does not suffer from over-smoothing or over-contrasting artifacts (Best viewed in color on a high-resolution display)}
\label{fig:RNI15}
\end{figure}

\subsection{Contributions}
\begin{itemize}
\item CNN based approaches for real image restoration (synthetic image denoising) producing state-of-the-art results using a first of its kind single-stage model.

\item  To the best of our knowledge, the first use of feature attention in restoration tasks, specifically in denoising, JPEG compression, and raindrop removal (although feature attention is used in super-resolution, however, our model is lightweight and efficient).

\item A modular network less affected by vanishing gradients~\cite{bengio1994vanishing}, thus enabling improved performance with an increasing number of modules.

\item Quantitative and qualitative experimental results on 11 real-image degradation datasets produce state-of-the-art results against more than 30 algorithms. Additionally, results on three synthetic noisy datasets are provided for denoising.

\item  We introduce a single network that can handle spatially variant noise, significant local artifacts (JPEG compression), pixel-by-pixel artifacts (raindrop removal), and super-resolution.  

\end{itemize}

This article is an extended version of our conference paper, RIDNet~\cite{anwar2019real}.  We have modified the RIDNet network and trained the network to show its performance against state-of-the-art deep networks on more applications, including JPEG, raindrop removal, and super-resolution. Furthermore, RIDNet is specific to real image denoising; therefore,  in this article, we termed the network as real restoration network \enquote{R$^2$Net} to accommodate broad applications under comparison.

\section{Related Work}
We present the literature for denoising, super-resolution, JPEG compression, and raindrop removal in the following sections.  

\subsection{Image denoising}
In this section, we present and discuss recent trends in image denoising.  Two notable denoising algorithms, NLM~\cite{Buades2005NLM} and BM3D~\cite{Dabov2007BM3D} use self-similar patches.  Due to their success, many subsequent variants were proposed, including SADCT~\cite{Foi2007SADCT},  SAPCA~\cite{Dabov2009BM3DSAPCA}, and NLB~\cite{Lebrun2013NLB} which seek self-similar patches in different transform domains. Dictionary-based methods~\cite{Elad2009ERD,Dong2011CSR} enforce sparsity by employing self-similar patches and learning over-complete dictionaries from clean images. Many algorithms \cite{Zoran2011EPLL,Xu2015PG-GMM} investigated the maximum likelihood algorithm to learn a statistical prior, \eg the Gaussian Mixture Model of natural patches or patch groups for patch restoration.  Furthermore, Levin \etal \cite{Levin2011Bounds} and  Chatterjee \etal \cite{Chatterjee2010IDD},  motivated external denoising~\cite{anwar2017category,anwar2017combined,luo2015adaptive} by showing that an image can be recovered with negligible error by selecting reference patches from a clean external database. However, all of the external algorithms are class-specific.

Recently, Schmidt \etal \cite{schmidt2014CSF} introduced a cascade of shrinkage fields (CSF) which integrated half-quadratic optimization and random-fields. Shrinkage aims to suppress smaller values (noise values) and learn mappings discriminatively. CSF assumes the data fidelity term to be quadratic and that it has a discrete Fourier transform-based closed-form solution.

Due to the popularity of convolutional neural networks (CNNs), image denoising algorithms~\cite{zhang2017DnCNN,zhang2017IrCNN,lefkimmiatis2017NLNet,Burger2012MLP,schmidt2014CSF,anwar2017chaining} have achieved a performance boost. Notable denoising neural networks, DnCNN \cite{zhang2017DnCNN}, and IrCNN \cite{zhang2017IrCNN} predict the residue present in the image instead of the denoised image itself as the input to the loss function. That is, comparing against ground truth noise instead of the original clean image. Both networks achieved better results despite having a simple architecture where repeated blocks of convolutional, batch normalization, and ReLU activations are used. Furthermore,  IrCNN \cite{zhang2017IrCNN} and DnCNN \cite{zhang2017DnCNN} are dependent on blindly predicted noise \ie without taking into account the underlying structures and textures of the noisy image.

Another essential image restoration framework is Trainable Nonlinear Reaction-Diffusion (TNRD) \cite{chen2017TNRD}, which uses a field-of-experts prior \cite{roth2009fields} into the deep neural network for a specific number of inference steps by extending the non-linear diffusion paradigm into a profoundly trainable set of parametrized linear filters and influence functions.  Although the results of TNRD are favorable, the model requires a significant amount of data to learn the parameters and influence functions as well as overall fine-tuning, hyper-parameter determination, and stage-wise training.  Similarly, non-local color net (NLNet) \cite{lefkimmiatis2017NLNet} was motivated by non-local self-similar (NSS) priors, which employ non-local self-similarity coupled with discriminative learning. NLNet improved upon the traditional methods; but, it lags in performance compared to most of the CNNs \cite{zhang2017IrCNN,zhang2017DnCNN} due to the adaptation of NSS priors, as it is unable to find the analogs for all the patches in the image. Recently,  Anwar~\etal introduced CIMM, a deep denoising CNN architecture, composed of identity mapping modules~\cite{anwar2017chaining}. The network learns features in cascaded identity modules using dilated kernels and uses self-ensemble to boost performance. CIMM improved upon all the previous CNN models~\cite{zhang2017DnCNN,jiao2017formresnet}.


\begin{figure*}
\begin{center}
\includegraphics[clip, trim=3.5cm 1.5cm 5.3cm 3cm, width=\textwidth]{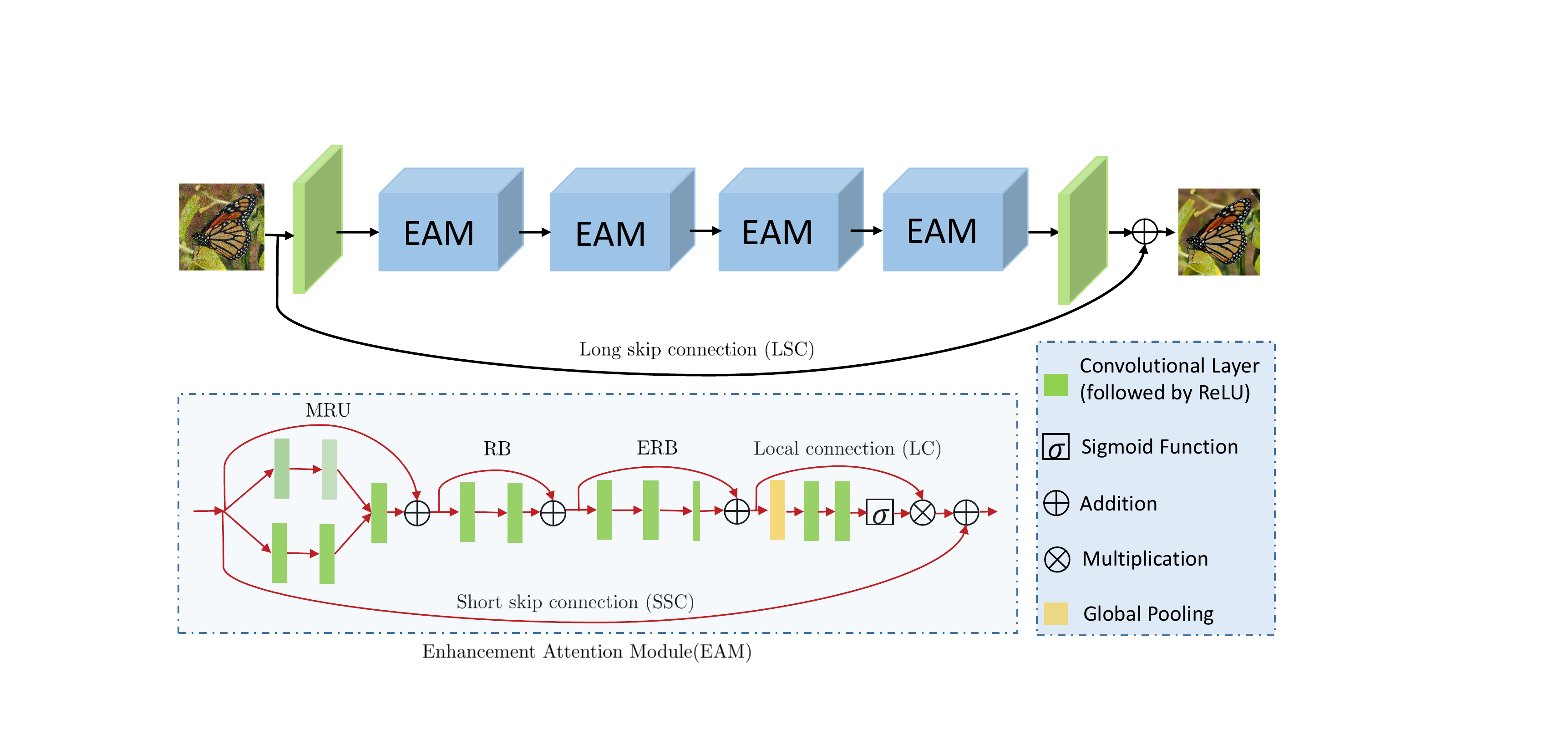}\\ 
\end{center}
\caption {The architecture of the proposed network. Different variants of green colors of the conv layers denote different dilations while the smaller width of the conv layer means the kernel is $1 \times 1$.  The second row shows the architecture of each EAM.}
\label{fig:net_architecture}
\end{figure*}

Recently, many algorithms focused on blind denoising on real-noisy images~\cite{plotz2018N3Net,guo2018CBDnet,brooks2019UPI}. The algorithms~\cite{zhang2017IrCNN,zhang2017DnCNN,jiao2017formresnet} benefitted from the modeling capacity of CNNs and have shown the ability to learn a single-blind denoising model; however, the denoising performance is limited, and the results are not satisfactory on real photographs.  Generally speaking, real-noisy image denoising is a two-step process: the first step involves noise estimation, whereas the second step addresses non-blind denoising. Noise clinic (NC)~\cite{lebrun2015NC} estimates the noise model dependent on signal and frequency followed by denoising the image using non-local Bayes (NLB). In comparison, Zhang~\etal~\cite{zhang2018ffdnet} proposed a non-blind Gaussian denoising network, termed FFDNet, that can produce satisfying results on some of the real-noisy images; however, it requires manual intervention to select a high value for high noise-level.

More recently, CBDNet~\cite{guo2018CBDnet} was used to train a blind denoising model for real photographs. CBDNet~\cite{guo2018CBDnet} is composed of two subnetworks: noise estimation and non-blind denoising.  CBDNet~\cite{guo2018CBDnet} also incorporated multiple losses, is engineered to be trained on real-synthetic noise and real-image noise, and enforces a higher noise standard deviation value for low noise images. Similarly, The methods of~\cite{guo2018CBDnet,zhang2018ffdnet} may require manual intervention to improve the results. In this paper, we present an end-to-end architecture that learns the noise characteristics and produces results on real noisy images without requiring separate subnets or manual intervention.

\subsection{Super-resolution}
In this section, we provide a chronological record of advancement in the area of deep super-resolution.  The initial focus of CNN models was to have a simple architecture with no skip connections. For example, SRCNN~\cite{dong2016SRCNNPAMI} with three convolutional layers and FSRCNN~\cite{dong2016FSRCNN} having eight convolutional layers utilizing shrinking and expansion of channels to make it run in real-time on a CPU.  Next, SRMD~\cite{zhang2018SRMDNF}, a linear network (resembling ~\cite{dong2016SRCNNPAMI,zhang2017IrCNN}), is able to handle multiple types of degradations. The input to the system is low-resolution images with the corresponding degradation maps. 

The introduction of skip-connections in deep networks made its way into super-resolution algorithms. Kim~\etal~\cite{kim2016VDSR}  employed a global skip connection to enforce residual learning, improving on the previous super-resolution methods. The same authors then developed a deep recursive structure (DRCN)~\cite{kim2016DRCN} sharing layer parameters, which reduced the number of parameters significantly, though, it lags behind VDSR~\cite{kim2016VDSR} in performance. Following that, to decrease the memory usage and computational complexity, Tai~\etal~\cite{tai2017DRRN} proposed a deep recursive residual network (DRRN) that utilizes basic skip-connections to implement residual learning along various convolutional blocks \ie multi-path architecture. 

The success of residual blocks motivated many super-resolution works. In the enhanced deep super-resolution (EDSR) network, Lim~\etal~\cite{lim2017EDSR} proposed to employ residual blocks and a global skip-connection rescaling each block output by a factor of 0.1 to avoid exploding gradients and significantly improving on all previous methods. More recently, Ahn~\etal~\cite{ahn2018CARN} proposed an efficient network, namely, the cascading residual network (CARN). The authors use cascading connections with a variant of residual blocks with three convolutional layers.   

To improve performance, several super-resolution works are driven by the success of dense-concatenation~\cite{huang2017densely}. For example, Tong~\etal~\cite{tong2017image} takes the output of all the previous convolutional layers in a block and fed into the subsequent one. Similarly, residual-dense network (RDN)~\cite{zhang2018RDN} learns relationships through dense-connections in the patches. Lately, Haris~\etal~\cite{haris2018DDBPN} trains a series of densely connected downsampling and upsampling layers (single block) with feedback and feed-forward mechanism.   

Recently, Zhang~\etal~\cite{zhang2018RCAN} introduced visual attention~\cite{mnih2014recurrent} in their RCAN network. In addition, the authors also employ a series of residual blocks and multi-level skip-connections in the network. Furthermore, Kim~\etal~\cite{kim2018ram}, in parallel to~\cite{zhang2018RCAN}, suggested a dual attention procedure, namely, SRRAM. The performance of SRRAM~\cite{kim2018ram} is, however, not on par with RCAN~\cite{zhang2018RCAN}. Recently, Anwar \& Barnes introduced densely connected residual units with laplacian attention~\cite{anwar2019densely} to advance super-resolution performance. Comparatively, our proposed SR method is lightweight and achieves competitive performance.

\subsection{Raindrop Removal}
Many papers deal with visibility enhancement, which removes haze, fog, and rain streaks; however, these algorithms do not apply to raindrop removal on a window or camera lens as the image formation models are different. 

To remove raindrops, Kurihata~\etal~\cite{kurihata2005rainy} proposes to apply PCA on the learned shapes of raindrops. During testing, the learned shapes are compared to raindrops, and the matching entities are removed. Due to the irregular shapes of raindrops, it is challenging to learn all the representative shapes; similarly, it is not easy to model transparent raindrops via PCA. Furthermore, there is a risk that local content is falsely detected and removed from the image, mainly due to similar appearance and structure. Roser \& Gieger~\cite{roser2009video} compare raindrops generated synthetically with real ones based on the assumption that the former are simply spherical while the latter ones are inclined spheres. Although the assumptions seem to enhance visibility, however, the method lacks generalization capability due to the random sizes and shapes of raindrops.  

To detect raindrops, Yamashita \etal~\cite{yamashita2005removal} employed stereo images using disparity measures. Neighborhood textures replaced the detected raindrops based on the assumption that the raindrop occludes a similar-looking background. Furthermore, Yamashita~\etal~\cite{yamashita2009noises} proposed a method relevant to~\cite{kurihata2005rainy}, where a sequence of images was used instead of stereo pairs. Similarly, You~\etal~\cite{you2015adherent} exploited a motion-based method for raindrop detection, while a video completion method is employed to remove it. The performance of these methods may be satisfactory in certain video scenes; however, a straightforward application to the case of a single image is not possible.  

Eigne~\etal~\cite{eigen2013restoring} and DeRain~\cite{qian2018DeRain} are the only methods specifically designed for this task. Eigne~\etal~\cite{eigen2013restoring} employs a shallow network with three layers, which works for sparse and small raindrops. However, it fails to remove dense or large drops, and the output of the network is not sharp. More recently, DeRain~\cite{qian2018DeRain} is proposed by Qian~\etal, which uses a GAN as the backbone with LSTMs, attention, and both global and local assessments. Finally, we also compare with a general method, namely, Pix2Pix~\cite{isola2017image}, that maps the input image to the output one by minimizing a loss function. On the other hand, we use attention to remove the raindrops without employing multiple backbone networks or loss functions.

\subsection{JPEG Compression}
JPEG algorithms fall into two categories: 1) deblocking-oriented and 2) restoration-oriented.  Deblocking-oriented approaches remove the blocking artifacts in the spatial domain by using adaptive filters~\cite{list2003adaptive,wang2013adaptive}. While in the frequency domain, transforms and thresholds are used at multiple scales to eliminate the artifacts, \eg the Pointwise Shape-Adaptive DCT (SA-DCT)~\cite{Foi2007SADCT}.  Although deblocking methods remove artifacts, it fails to generate sharp edges and produce smooth textures. 

On the other hand, the restoration based approach considers the compression as a form of distortion, including the sparse-coding-based method~\cite{jung2012Sparse}, the Regression Tree Fields based method (RTF)~\cite{jancsary2012RTF}, and CNN-based methods~\cite{dong2015ARCNN,zhang2017DnCNN} \etc Recently, Artifacts Reduction Convolutional Neural Networks (AR-CNN) ~\cite{dong2015ARCNN}, inspired by and with similar architecture as SRCNN~\cite{dong2016SRCNNPAMI} except having more feature layers.
AR-CNN~\cite{dong2015ARCNN} lags behind DnCNN~\cite{zhang2017DnCNN} in performance.

Li~\etal~\cite{li2014LD} formulated the JPEG artifacts removal as an ill-posed inverse image decomposition problem and solved it through optimization. Similarly, Fan~\etal~\cite{fan2018LPIO} proposed a decouple framework for image restoration tasks using several image operators. More recently, DCSC~\cite{fu2019DCSC} is specifically proposed to tackle the JPEG compression artifacts via a deep learning network employing sparse coding. Contrary to the mentioned networks, our method is useful in suppressing artifacts and preserving edges and sharp details via feature attention, merge-and-run units, and enhanced residual modules.

\section{CNN restorer}
\subsection{Network Architecture}
Our model is composed of three main modules, \ie feature extraction, feature learning residual on the residual, and reconstruction, as shown in Figure \ref{fig:net_architecture}.  Let us consider $x$ a degraded input image, and $\hat{y}$ the restored output image.  Our feature extraction module is composed of only one convolutional layer to extract initial features $f_0$ from the noisy input:
\begin{ceqn}
\begin{equation}
f_0 = M_e(x),
\label{eq:extraction}    
\end{equation}
\end{ceqn}
where $M_e(\cdot)$ performs convolution on the noisy input image. Next, $f_0$ is passed on to the feature learning residual on the residual module, termed $M_{fl}$, 
\begin{ceqn}
\begin{equation}
f_r = M_{fl}(f_0),
\label{eq:fl}
\end{equation}  
\end{ceqn}
where $f_r$ are the learned features and $M_{fl}(\cdot)$ is the main feature learning residual on the residual component, composed of enhancement attention modules (EAM) that are cascaded together as shown in Figure~\ref{fig:net_architecture}. Our network has a small depth but provides a wide receptive field through kernel dilation in each of the first two branches of convolutions in the EAM. The output features of the final layer are fed to the reconstruction module, which is again composed of one convolutional layer.

\begin{ceqn}
\begin{equation}
\hat{y} = M_r(f_r),
\label{eq:r}
\end{equation}
\end{ceqn}
where $M_r(\cdot)$ denotes the reconstruction layer. 

There are several choices available for the loss function to optimize such as  $\ell_2$~\cite{zhang2017DnCNN,zhang2017IrCNN,anwar2017chaining}, perceptual loss~\cite{jiao2017formresnet,guo2018CBDnet}, total variation loss~\cite{jiao2017formresnet} and asymmetric loss~\cite{guo2018CBDnet}. Some networks~\cite{jiao2017formresnet,guo2018CBDnet} make use of more than one loss to optimize the model. Contrary to  earlier networks, we only employ one loss, \ie $\ell_1$. Now, given a batch of $N$ training pairs, $\{x_i, y_i\}_{i=1}^N$, where $x$ is the noisy input and $y$ is the ground truth, the aim is to minimize the $\ell_1$ loss function 
\begin{ceqn}
\begin{equation}
L(\mathcal{W}) = \frac{1}{N} \sum_{i=1}^N||\text{R$^2$Net}(x_i) - y_i||_1,
\label{eq:l1_loss}
\end{equation}
\end{ceqn}
where R$^2$Net($\cdot$) is our network, and $\mathcal{W}$ denotes the set of all the network parameters learned. Our feature extraction $M_e$ and reconstruction module $M_r$ resemble the previous algorithms~\cite{dong2016SRCNNPAMI,anwar2017chaining}. We now focus on the feature learning residual on the residual block and feature attention.

\subsection{Feature Learning Residual on the Residual}
\label{ss:fl_rir}
This section provides more detail on the enhancement attention modules that use a Residual on the Residual structure with a local skip and short skip connections.  Each EAM is further composed of $D$ blocks, followed by feature attention. Thanks to the residual on the residual architecture, very deep networks are now possible that improve denoising performance; however, we restrict our model to four EAM modules only. The first part of EAM covers the full receptive field of input features, followed by learning on the features; then, the features are compressed for speed, and finally, a feature attention module enhances the weights of essential features from the maps. The first part of EAM is realized using a novel merge-and-run unit (MRU), as shown in the second row of Figure~\ref{fig:net_architecture}. The input features are branched and passed through two parallel dilated convolutions, then concatenated and passed through another convolution. Next, the features are learned using a residual block (RB) of two convolutions, while compression is achieved by an enhanced residual block (ERB) of three convolutional layers. The last layer of ERB flattens the features by applying a $1 \times 1$ kernel. Finally, the output of the feature attention unit is added to the input of EAM.

In image recognition, residual blocks \cite{he2016deep} are often stacked together to construct a network of more than 1k layers. Similarly, in image superresolution, EDSR \cite{lim2017EDSR} stacked the residual blocks and used long skip connections (LSC) to form a very deep network. However, to date, very deep networks have not been investigated for denoising. Motivated by the success of \cite{zhang2018RCAN}, we introduce the residual on the residual as a basic module for our network to construct deeper systems. Now consider the m-th module of the EAM is given as
\begin{ceqn}
\begin{equation}
f_m = EAM_m(EAM{m-1}(\cdots (M_0(f_0)) \cdots)),
\label{eq:rir_block}
\end{equation}
\end{ceqn}
where $f_m$ is the output of the $EAM_m$ feature learning module, in other words $f_m = EAM_m(f_{m-1})$. The output of each EAM is added to the input of the group as $f_m = f_m+f_{m-1}$.  The learned features \ie $M_{fl}(\mathcal{W}_{w,b})$ are passed to the reconstruction layer to output the same number of channels as the input of the network. Furthermore, we use a long skip connection to add the input image to the final network output as
\begin{ceqn}
\begin{equation}
\hat{y}  = x  + M_c(M_{fl}(\mathcal{W}_{w,b})),
\label{eq:residual_block}
\end{equation}
\end{ceqn}
where $\mathcal{W}_{w,b}$ are the weights and biases learned in the group. This addition, \ie LSC eases the flow of information across groups and helps learning the residual (degradation) rather than the image. This technique helps in faster learning as compared to learning the original image thanks to the sparse representation of the degradation. 

\subsubsection{Feature Attention}
\begin{figure}[t]
\begin{center}
\includegraphics[clip, trim=5cm 8cm 5cm 6cm, width=0.5\textwidth]{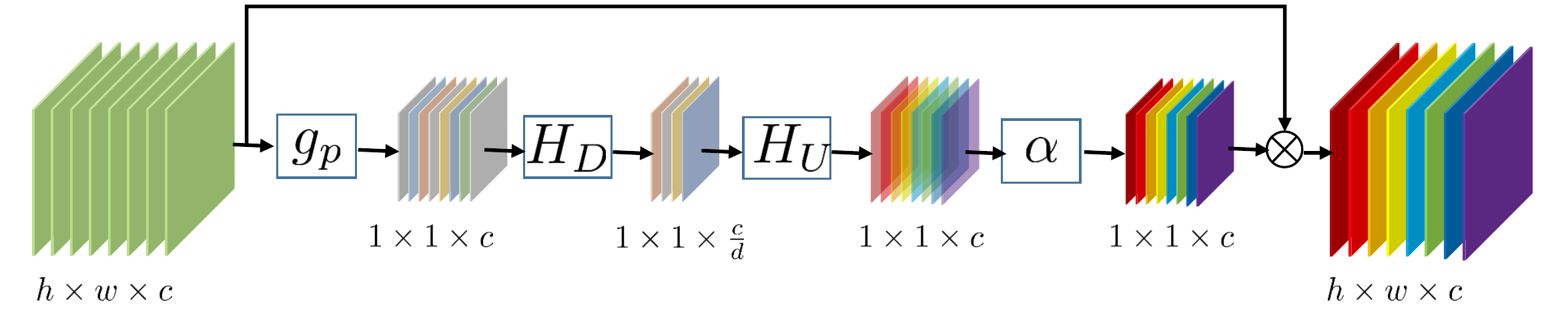}\\ 
\end{center}
\caption {The feature attention mechanism for selecting the essential features.}
\label{fig:net_eam}
\end{figure}

\begin{table*}[t]
\caption{ Investigation of skip connections and feature attention. The best result in PSNR (dB) on values on BSD68~\cite{roth2009fields} in 2$\times10^{5}$ iterations is presented.}
\centering
\begin{tabular}{l||c|c|c|c|c|c|c|c|c}
\hline\hline
 Long skip connection (LSC)     &      & \ch   & 	  &  \ch &       &        &       &  \ch   &\ch \\ 
  Short skip connection (SSC)   &      &  	   & \ch  &  \ch &       &        & \ch   &  \ch   &\ch \\  
  Long connection (LC)          &      &  	   & 	  &      &       & \ch          & \ch   &        &\ch \\
 Feature attention (FA)         &      &  	   & 	  &      &  \ch  & \ch       & \ch   & \ch    &\ch \\  \hline
 PSNR (in dB)  &28.45 & 28.77 &28.81 &28.86 & 28.52 &28.85  & 28.86 & 28.90 &28.96 \\  \hline \hline
\end{tabular}
\label{table:ablation}
\end{table*}

This section provides information about the feature attention mechanism. Attention \cite{xu2015show} has been around for some time; however, it has not been employed in image denoising. Channel features in image denoising methods are treated equally, which is not appropriate for many cases.  To exploit and learn the image's critical content, we focus attention on the relationship between the channel features; hence, the name: feature attention (see Figure~\ref{fig:net_eam}).

An important question here is how to generate attention differently for each channel-wise feature. Images generally can be considered as having low-frequency regions (smooth or flat areas), and high-frequency regions (\eg, lines edges, and texture). As convolutional layers exploit local information only and are unable to utilize global contextual information, we first employ global average pooling to express the statistics denoting the whole image, other options for aggregation of the features can also be explored to represent the image descriptor. Let $f_c$ be the output features of the last convolutional layer having $c$ feature maps of size $h \times w$; global average pooling will reduce the size from $h \times w \times c$ to $1 \times 1 \times c$ as:
\begin{ceqn}
\begin{equation}
g_p = \frac{1}{h \times w} \sum_{i=1}^h \sum_{i=1}^w f_c(i,j),
\label{eq:GAP}
\end{equation}
\end{ceqn}
where $f_c(i,j)$ is the feature value at position $(i,j)$ in the feature maps. 

Furthermore, as investigated in \cite{hu2018squeeze}, we propose a self-gating mechanism to capture the channel dependencies from the descriptor retrieved by global average pooling. According to \cite{hu2018squeeze}, the mentioned mechanism must learn the non-linear synergies between channels as well as mutually-exclusive relationships. Here, we employ soft-shrinkage and sigmoid functions to implement the gating mechanism. Let us consider $\delta$, and $\alpha$ are the soft-shrinkage and sigmoid operators, respectively. Then the gating mechanism is
\begin{ceqn}
\begin{equation}
r_c =  \alpha(H_U(\delta(H_D(g_p)))),
\label{eq:gating}
\end{equation}
\end{ceqn}
where $H_D$ and $H_U$ are the channel reduction and channel upsampling operators, respectively. The global pooling layer's output $g_p$ is convolved with a downsampling Conv layer, activated by the soft-shrinkage function. To differentiate the channel features, the output is then fed into an upsampling Conv layer followed by sigmoid activation.   Moreover, to compute the statistics, the output of the sigmoid ($r_c$) is adaptively rescaled by the input $f_c$ of the channel features as
\begin{ceqn}
\begin{equation}
\hat{f}_c = r_c\times f_c
\label{eq:rescale_CA}
\end{equation}
\end{ceqn}
\subsection{Implementation}
Our proposed model contains four EAM blocks. The kernel size for each convolutional layer is set to $3\times3$, except the last Conv layer in the enhanced residual block and those of the features attention units, where the kernel size is $1 \times 1$. Zero padding is used for $3 \times 3$ to achieve the same size outputs feature maps. The number of channels for each convolutional layer is fixed at 64, except for feature attention downscaling. A factor of 16 reduces these Conv layers, hence having only four feature maps. The final convolutional layer either outputs three or one feature maps depending on the input. As for running time, our method takes about 0.2 seconds to process a $512 \times 512$ image.

\section{Experiments}

\subsection{Training settings}
To generate noisy synthetic images, we employ BSD500~\cite{Martin2001BSD}, DIV2K~\cite{agustsson2017ntire}, and MIT-Adobe FiveK~\cite{bychkovsky2011learning}, resulting in 4k images while for real noisy images, we use cropped patches of $512 \times 512$ from SSID~\cite{abdelhamed2018high}, Poly~\cite{xu2018real}, and RENOIR~\cite{anaya2018renoir}. Data augmentation is performed on training images, which includes random rotations of 90$^{\circ}$, 180$^{\circ}$, 270$^{\circ}$ and flipping horizontally. In each training batch, 32 patches are extracted as inputs with a size of $80 \times 80$.  Adam~\cite{kingma2014adam} is used as the optimizer with default parameters. The learning rate is initially set to $10^{-4}$ and then halved after $10^{5}$ iterations. The network is implemented in the Pytorch~\cite{paszke2017automatic} framework and trained with an Nvidia Tesla V100 GPU. Furthermore, we use PSNR as the evaluation metric.

\subsection{Ablation Studies}

\subsubsection{Influence of the skip connections}
Skip connections play a crucial role in our network. Here, we demonstrate the effectiveness of the skip connections. Our model is composed of three basic types of connections, which include long skip connection (LSC),  short skip connections (SSC), and local connections (LC). Table~\ref{table:ablation} shows the average PSNR for the BSD68~\cite{roth2009fields} dataset. The highest performance is obtained when all the skip connections are available while the performance is lower when any connection is absent. We also observed that increasing the network's depth in the absence of skip connections does not benefit performance.

\begin{table*}
\caption{The similarity between the denoised and the clean images of BSD68 dataset~\cite{roth2009fields} for our method and competing measured in terms of average PSNR for $\sigma$=15, 25, and 50 on grayscale images.}
\centering
\begin{tabular}{c|c|c|c|c|c|c|c|c|c|c}
\hline\hline
 Noise & \multicolumn{9}{c}{Methods} \\ 
 Level& BM3D    & WNNM  	& EPLL	  & TNRD   & DenoiseNet & DnCNN   & IrCNN   & NLNet     &FFDNet	& Ours	  \\\hline 
 15    & 31.08   & 31.32 	& 31.19	  & 31.42  & 31.44 	    & 31.73   &31.63    & 31.52	    &31.63	&\textbf{31.81} \\ 
 25    & 28.57   & 28.83	& 28.68	  & 28.92  & 29.04 	    & 29.23   &29.15    & 29.03 	&29.23	&\textbf{29.34} \\  
 50    & 25.62   & 25.83	& 25.67	  & 26.01  & 26.06  	& 26.23   &26.19    & 26.07	    &26.29	&\textbf{26.40} \\ \hline \hline
\end{tabular}
\label{table:BSD68_grayscale}
\end{table*}

\begin{table*}
\caption{Performance comparison between our network and existing state-of-the-art algorithms on the color version of the BSD68 dataset~\cite{roth2009fields}.}
\centering
\begin{tabular}{c|c|c|c|c|c|c|c|c}
\hline \hline
Noise  & \multicolumn{7}{c}{Methods} \\ 
Levels & CBM3D~\cite{dabov2007CBM3D}       & MLP~\cite{Burger2012MLP}    & TNRD~\cite{chen2017TNRD}  & DnCNN~\cite{zhang2017DnCNN}  & IrCNN~\cite{zhang2017IrCNN}  & CNLNet~\cite{lefkimmiatis2017NLNet} &FFDNet~\cite{zhang2018ffdnet}& Ours   \\ \hline
 15    &  33.50      & -      & 31.37 & 33.89   					& 33.86 & 33.69	&33.87 & \textbf{34.01}\\  
 25    &  30.69      & 28.92  & 28.88 & 31.33   	& 31.16 & 30.96 &31.21 & \textbf{31.37} \\  
 50    &  27.37      & 26.00  & 25.94 & 27.97   					& 27.86 & 27.64	&27.96 & \textbf{28.14}\\ \hline  \hline        
\end{tabular}
\label{table:BSD68_color}
\end{table*}

\begin{table}
\caption{The quantitative comparison between denoising algorithms on 12 classical images, (in terms of PSNR). The best results are highlighted as bold.}
\centering
\begin{tabular}{|l||ccc|}\hline 
Methods    &   $\sigma$ = 15 & $\sigma$ = 25    &   $\sigma$ = 50 \\ \hline \hline
BM3D~\cite{Dabov2007BM3D}    & 32.37 &   29.97  &   26.72 \\
WNNM~\cite{Gu2014WNN}        & 32.70 &   30.26  &   27.05 \\
EPLL~\cite{Zoran2011EPLL}    & 32.14 &   29.69  &   26.47 \\
MLP~\cite{Burger2012MLP}     & -     &   30.03  &   26.78 \\
CSF~\cite{schmidt2014CSF}    & 32.32 &   29.84  &    -    \\
TNRD~\cite{chen2017TNRD}     & 32.50 &   30.06  &   26.81 \\
DnCNN~\cite{zhang2017DnCNN}  & 32.86 &   30.44  &   27.18 \\
IrCNN ~\cite{zhang2017IrCNN} & 32.77 &   30.38  &   27.14 \\ 
FFDNet~\cite{zhang2018ffdnet}& 32.75 &   30.43  &   27.32 \\
Ours                         & \textbf{32.91} &   \textbf{30.60}  &   \textbf{27.43} \\ \hline \hline
\end{tabular}
\label{table:classical}
\end{table}

\subsubsection{Feature-attention}
Another important aspect of our network is feature attention. Table~\ref{table:ablation} compares the PSNR values of the networks with and without feature attention. The results support our claim about the benefit of using feature attention. Since the inception of DnCNN~\cite{zhang2017DnCNN}, the CNN models have matured, and further performance improvement requires the careful design of blocks and rescaling the feature maps. The two mentioned characteristics are present in our model in the form of feature-attention and skip connections.

\subsection{Denoising Comparisons}
We evaluate our algorithm using the Peak Signal-to-Noise Ratio (PSNR) index as the error metric and compare against many state-of-the-art competitive algorithms which include traditional methods \ie CBM3D~\cite{Dabov2007BM3D}, WNNM~\cite{Gu2014WNN},  EPLL~\cite{Zoran2011EPLL}, CSF~\cite{schmidt2014CSF} and CNN-based denoisers \ie MLP~\cite{Burger2012MLP},  TNRD~\cite{chen2017TNRD}, DnCNN~\cite{zhang2017DnCNN}, IrCNN~\cite{zhang2017IrCNN}, CNLNet~\cite{lefkimmiatis2017NLNet}, FFDNet~\cite{zhang2018ffdnet} and CBDNet~\cite{guo2018CBDnet}. To be fair in comparison, we use the default setting of the traditional methods provided by the corresponding authors.

\subsubsection{Denoising Test Datasets}
In the experiments, we test four noisy real-world datasets \ie RNI15~\cite{lebrun2015NC}, DND~\cite{plotz2017benchmarking}, Nam~\cite{nam2016holistic} and SSID~\cite{abdelhamed2018high}. Furthermore, we prepare three synthetic noisy datasets from the widely used 12 classical images, BSD68~\cite{roth2009fields} color and gray 68 images for testing. We corrupt the clean images by additive white Gaussian noise using noise sigma of 15, 25, and 50 standard deviations.

\begin{itemize}
\item Classical images: The denoising comparisons would be incomplete without testing on the traditional images. Here, we use 12 classical images for testing.

\item BSD68~\cite{roth2009fields} is composed of 68 images grayscale (CBSD68 is the same but with color images). The ground-truth images are available as the degraded dataset is synthetically created.

\item RNI15~\cite{lebrun2015NC} provides 15 real-world noisy images. Unfortunately, the clean images are not given for this dataset; therefore, only the qualitative comparison is presented.

\item Nam~\cite{nam2016holistic} comprises 11 static scenes and the corresponding noise-free images obtained by the mean of 500 noisy images of the same scene. The size of the images is enormous; hence, we cropped the images in $512 \times 512$ patches and randomly selected 110 from those for testing.

\item DnD is recently proposed by Plotz~\etal~\cite{plotz2017benchmarking}, which initially contains 50 pairs of real-world noisy and noise-free scenes. The scenes are further cropped into patches of size  $512 \times 512$ by the dataset providers, which resulted in 1000 smaller images. The near noise-free images are not publicly available, and the results (PSNR/SSIM) can only be obtained through the online system introduced by~\cite{plotz2017benchmarking}.

\item SSID~\cite{abdelhamed2018high} (Smartphone Image Denoising Dataset) is recently introduced.  The authors have collected 30k real noisy images and their corresponding clean images; however, only 320 images are released for training and 1280 images pairs for validation, as testing images are not released yet. We use the validation images for testing our algorithm and the competitive methods.
\end{itemize}

\begin{figure}
\begin{center}
\begin{tabular}{c@{ }  c@{ } c}
\includegraphics[trim={1.5cm 5.15cm  1.5cm  5cm },clip,width=.145\textwidth,valign=t]{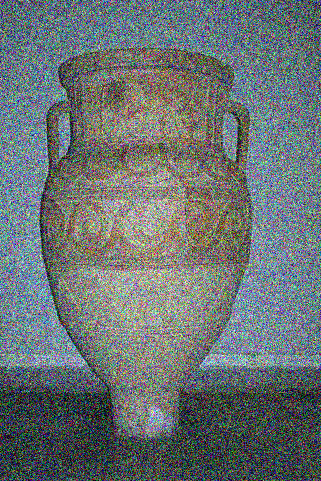}&   
\includegraphics[trim={1.5cm 5.15cm  1.5cm  5cm },clip,width=.145\textwidth,valign=t]{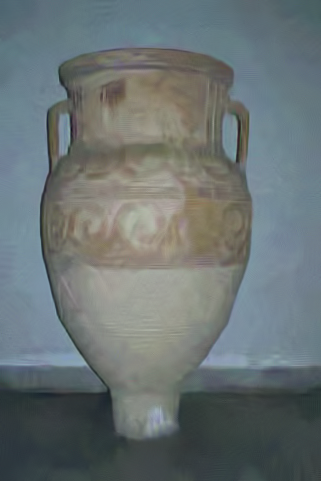}& 
\includegraphics[trim={1.5cm 5.15cm  1.5cm  5cm},clip,width=.145\textwidth,valign=t]{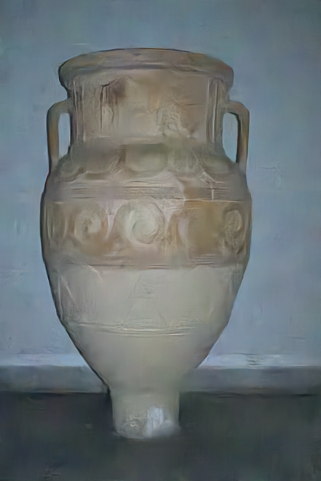}\\
& 31.68dB& 32.21dB\\
   Noisy & BM3D~\cite{dabov2007CBM3D}    & IRCNN~\cite{zhang2017IrCNN} \\

\includegraphics[trim={1.5cm 5.15cm  1.5cm  5cm },clip,width=.145\textwidth,valign=t]{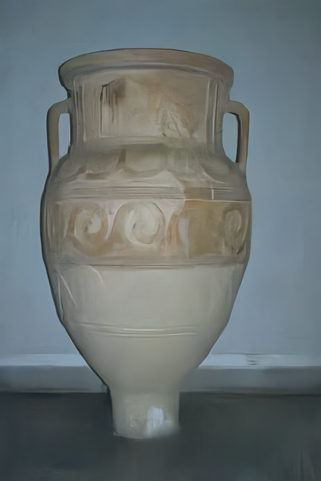}&
 \includegraphics[trim={1.5cm 5.15cm  1.5cm  5cm },clip,width=.145\textwidth,valign=t]{images/BSD68/227092_Ours}&
 \includegraphics[trim={1.5cm 5.15cm  1.5cm  5cm },clip,width=.145\textwidth,valign=t]{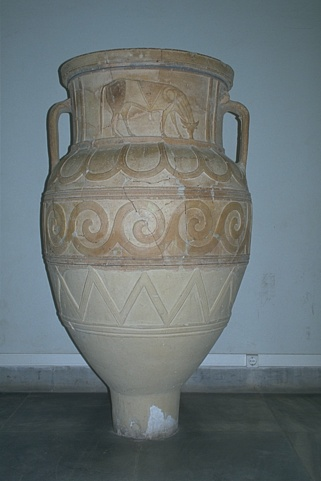}\\
 32.33dB& 32.84dB & \\
   DnCNN~\cite{zhang2017DnCNN} & Ours    & GT \\

\end{tabular}
\end{center}
\caption{Denoising performance of our R$2$Net versus state-of-the-art  methods on a color images from~\cite{roth2009fields} for $\sigma_n = 50$}
\label{fig:BSD68_color}
\end{figure}

\subsubsection{Classical noisy images}

\begin{figure*}[t]
\begin{center}
\begin{tabular}[b]{c@{ } c@{ }  c@{ } c@{ } c@{ } c@{ }	}
    \multirow{4}{*}{\includegraphics[width=.314\textwidth,valign=t]{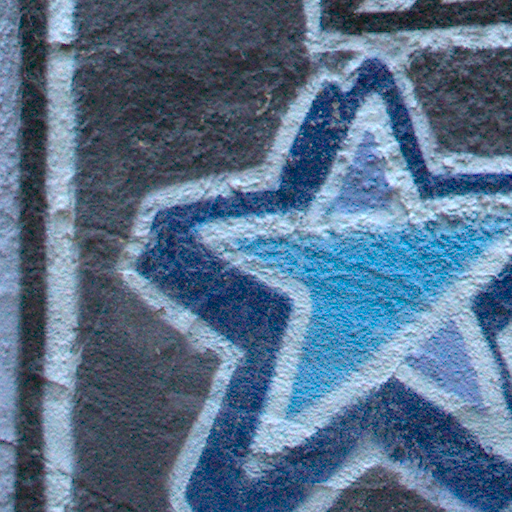}} &  
    \includegraphics[trim={3cm 3cm  3cm  3cm },clip,width=.133\textwidth,valign=t]{images/DnD/3_nosiy}&
  	\includegraphics[trim={3cm 3cm  3cm  3cm },clip,width=.13\textwidth,valign=t]{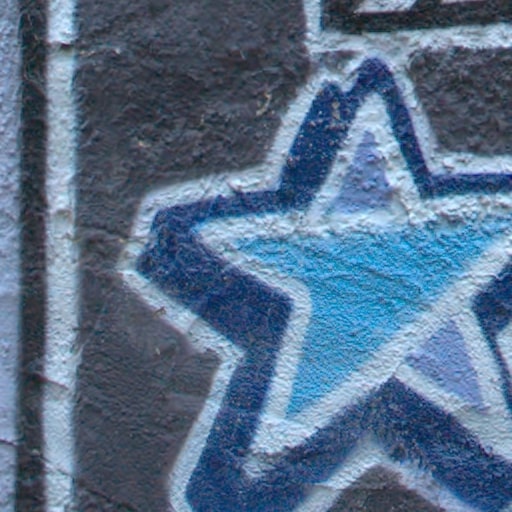}&   
    \includegraphics[trim={3cm 3cm  3cm  3cm },clip,width=.133\textwidth,valign=t]{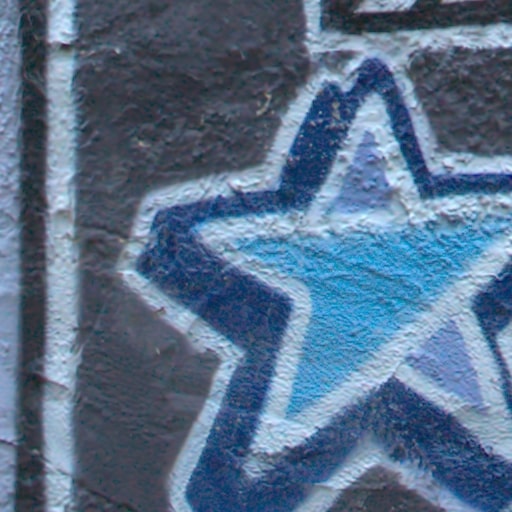}&
      	\includegraphics[trim={3cm 3cm  3cm  3cm },clip,width=.133\textwidth,valign=t]{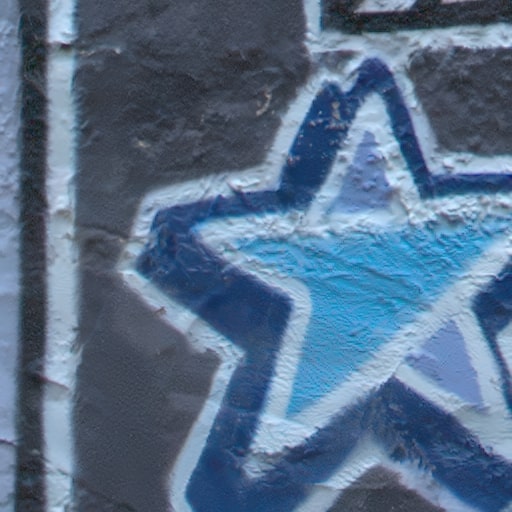}&
	\includegraphics[trim={3cm 3cm  3cm  3cm },clip,width=.133\textwidth,valign=t]{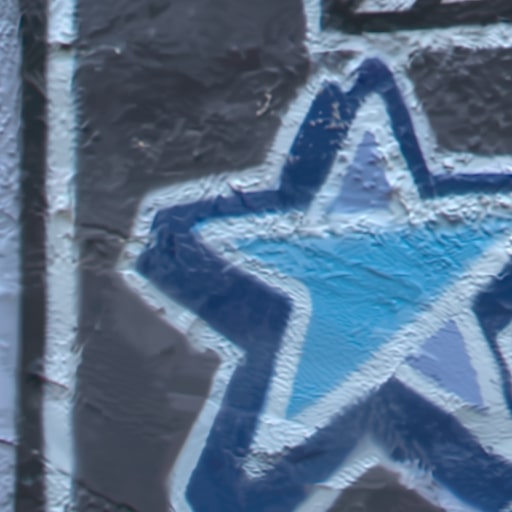}   
\\
    &       &30.896dB  & 29.98dB & 30.73dB & 29.42dB   \\
    & Noisy &CBM3D~\cite{Dabov2007BM3D}  & WNNM~\cite{Gu2014WNN}    & NC~\cite{lebrun2015NC}      & TWSC~\cite{xu2018TWSC} \\

    &
    \includegraphics[trim={3cm 3cm  3cm  3cm },clip,width=.133\textwidth,valign=t]{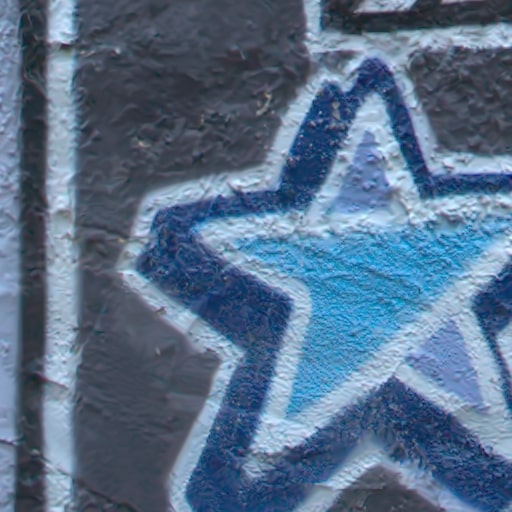}&
    \includegraphics[trim={3cm 3cm  3cm  3cm },clip,width=.133\textwidth,valign=t]{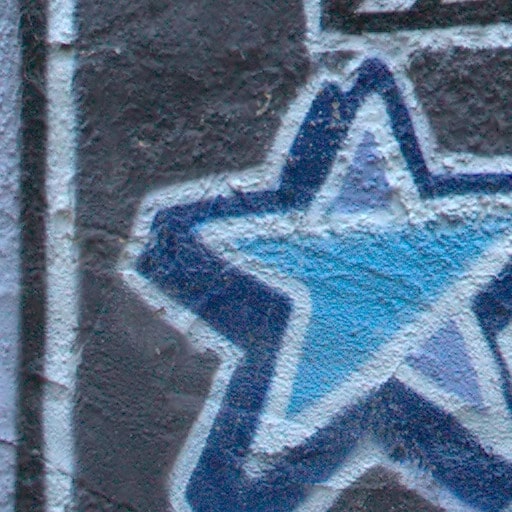}&
    \includegraphics[trim={3cm 3cm  3cm  3cm },clip,width=.133\textwidth,valign=t]{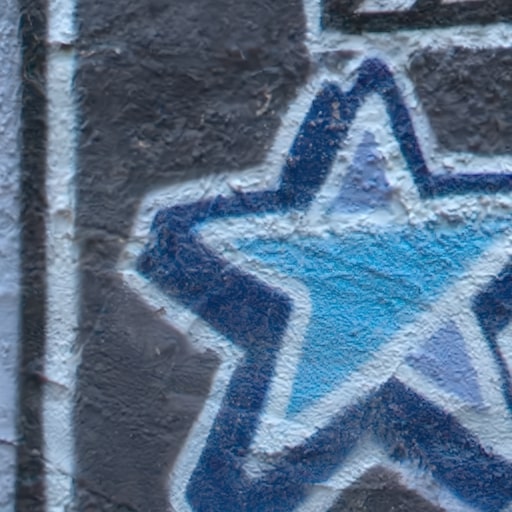}&  
    \includegraphics[trim={3cm 3cm  3cm  3cm },clip,width=.133\textwidth,valign=t]{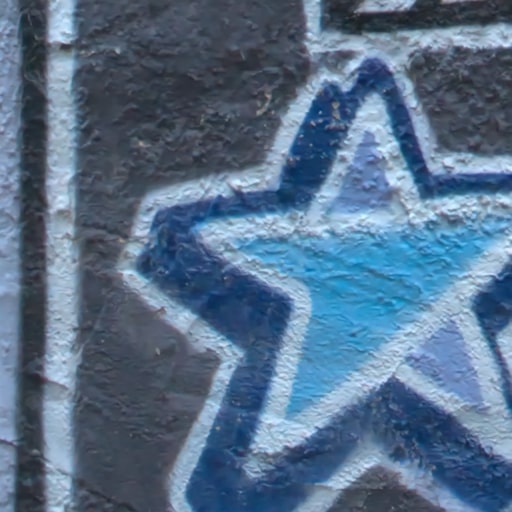}&   

     \includegraphics[trim={3cm 3cm  3cm  3cm },clip,width=.133\textwidth,valign=t]{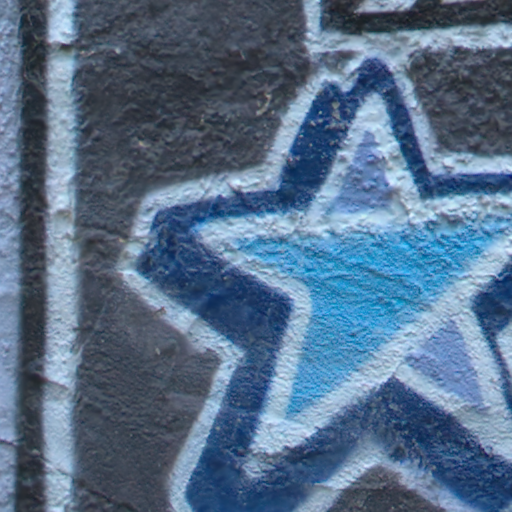}\\

     & 30.88dB & 28.43dB & 31.37dB & 31.06dB & \textbf{32.31dB}  \\
           Noisy Image  & MCWNNM~\cite{xu2017MCW}    & NI~\cite{NeatI}     &FFDNet~\cite{zhang2018ffdnet}       & CBDNet~\cite{guo2018CBDnet}     & R$^2$Net (Ours) \\
    
\end{tabular}
\end{center}
\caption{A real noisy example from DND dataset~\cite{plotz2017benchmarking} for comparison of our method against the state-of-the-art algorithms.}
\label{fig:DnD}
\end{figure*}
In this subsection, we evaluate our model on the noisy grayscale images corrupted by spatially invariant additive white Gaussian noise. We compare against nonlocal self-similarity representative models \ie BM3D~\cite{Dabov2007BM3D} and WNNM~\cite{Gu2014WNN}, learning based methods \ie EPLL, TNRD~\cite{chen2017TNRD}, MLP~\cite{Burger2012MLP}, DnCNN~\cite{zhang2017DnCNN}, IrCNN~\cite{zhang2017IrCNN}, and CSF~\cite{schmidt2014CSF}.  

\vspace{3mm}
\noindent
\textbf{SET12}:
In Table~\ref{table:classical}, we present the PSNR values on Set12. Our method outperforms all the competitive algorithms for all noise levels; this may be due to the larger receptive field in the merge-and-run unit as well as better network modeling capacity.  

\vspace{3mm}
\noindent
\textbf{BSD68}: We show the performance of our algorithm against competing methods on BSD68~\cite{roth2009fields} in Table~\ref{table:BSD68_grayscale}.  It is to be remembered here that BSD68~\cite{roth2009fields} and BSD500~\cite{Martin2001BSD} are two disjoint sets. Our method shows improvement over all the competitive algorithms for all noise levels. The increase in PSNR proves the superior network design and better feature learning for denoising tasks. It should be noted here that even a marginal improvement on the synthetic noisy datasets is difficult as according to the Levin~\etal~\cite{Levin2011Bounds} and Chatterjee~\cite{Chatterjee2010IDD} the synthetic denoising has already achieved optimal limits.

\vspace{3mm}
\noindent
\textbf{Color noisy images}: Next, for noisy color image denoising, we keep all the parameters of the network similar to the grayscale model, except the first and last layer are changed to input and output three channels rather than one. Figure~\ref{fig:BSD68_color} presents the visual comparison, and Table~\ref{table:BSD68_color} reports the PSNR numbers between our methods and the alternative algorithms. Our algorithm consistently outperforms all the other techniques published in Table~\ref{table:BSD68_color} for CBSD68 dataset~\cite{roth2009fields}.  Similarly, our network produces the best perceptual quality images, as shown in Figure~\ref{fig:BSD68_color}. A closer inspection of the vase reveals that our network generates textures closest to the ground-truth with fewer artifacts and more details.

\subsubsection{Real-World noisy images}

To assess the practicality of our model, we employ a real noise dataset.  The evaluation is difficult because of the unknown level of noise, the various noise sources such as shot noise, quantization noise \etc, imaging pipeline \ie image resizing, lossy compression \etc Furthermore, the noise is spatially variant (non-Gaussian) and also signal-dependent; hence, the assumption that noise is spatially invariant, employed by many algorithms does not hold for real image noise. Therefore, real-noisy images evaluation determines the success of the algorithms in real-world applications.

\begin{table}[t]
\caption{The Mean PSNR and SSIM denoising results of state-of-the-art algorithms evaluated on the DnD sRGB images~\cite{plotz2017benchmarking}}
\centering
\begin{tabular}{l||ccc}
\hline \hline
Method      &Blind &PSNR &SSIM \\ \hline \hline
CDnCNNB~\cite{zhang2017DnCNN}  	&\checkmark 		&32.43 &0.7900\\
EPLL~\cite{Zoran2011EPLL} 		& 	&33.51 &0.8244\\
TNRD~\cite{chen2017TNRD}  		& 	&33.65 &0.8306\\
NCSR~\cite{dong2012NCSR}  		& 	&34.05 &0.8351\\
MLP~\cite{Burger2012MLP}  		& 	&34.23 &0.8331\\
FFDNet~\cite{zhang2018ffdnet}  	&	&34.40 &0.8474\\
BM3D~\cite{Dabov2007BM3D}  		& 	&34.51 &0.8507\\
FoE~\cite{roth2009fields}  		& 	&34.62 &0.8845\\
WNNM~\cite{Gu2014WNN}  		    & 	&34.67 &0.8646\\
NC~\cite{lebrun2015NC}	        &\checkmark 	&35.43	&0.8841	\\
NI~\cite{NeatI}	    	        &\checkmark 	&35.11  &0.8778 \\
CIMM~\cite{anwar2017chaining}  	& 	&36.04 &0.9136\\
KSVD~\cite{aharon2006ksvd}  	& 	&36.49 &0.8978\\
MCWNNM~\cite{xu2017MCW}  		& 	&37.38 &0.9294\\
TWSC~\cite{xu2018TWSC}          &	&37.96 &0.9416\\
FFDNet+~\cite{zhang2018ffdnet}	&  &37.61 &0.9415\\
CBDNet~\cite{guo2018CBDnet}		&\checkmark  &38.06 &0.9421\\
R$^2$Net (Ours)		            &\checkmark  &\textbf{39.23} & \textbf{0.9526}\\ \hline \hline
                
\end{tabular}
\label{table:DnD}
\end{table}

\noindent
\textbf{DnD}: Table~\ref{table:DnD} presents the quantitative results (PSNR/SSIM) on the sRGB data for competitive algorithms and our method obtained from the online DnD benchmark website available publicly. The blind Gaussian denoiser DnCNN~\cite{zhang2017DnCNN} performs inefficiently and is unable to achieve better results than BM3D and WNNM due to the poor generalization of the noise during training.  Similarly, the non-blind Gaussian traditional denoisers are able to report limited performance, although the noise standard-deviation is provided. This may be due to the fact that these denoisers~\cite{Dabov2007BM3D,Gu2014WNN,Zoran2011EPLL} are tailored for AWGN only, and real-noise is different in characteristics to synthetic noise. Incorporating feature attention and capturing the appropriate characteristics of the noise through a novel module means our algorithm leads by a large margin \ie 1.17dB PSNR compared to the second performing method, CBDNet~\cite{guo2018CBDnet}. Furthermore, our algorithm only employs real-noisy images for training using only $\ell_1$ loss while CBDNet~\cite{guo2018CBDnet} uses many techniques such as multiple losses (\ie total variation, $\ell_2$ and asymmetric learning) and both real-noise as well as synthetically generated real-noise. As reported by the author of CBDNet~\cite{guo2018CBDnet}, it is able to achieve 37.72 dB with real-noise images only. Noise Clinic (NC)~\cite{lebrun2015NC} and Neat Image (NI)~\cite{NeatI} are the other two state-of-the-art blind denoisers other than~\cite{guo2018CBDnet}. NI~\cite{NeatI} is commercially available as a part of Photoshop and Corel PaintShop. Our network is able to achieve 3.82dB and 4.14dB more PSNR from NC~\cite{lebrun2015NC} and NI~\cite{NeatI}, respectively. 

Next, we visually compare our method's result with the competing methods on the denoised images provided by the online system of Plotz~\etal~\cite{plotz2017benchmarking} in Figure~\ref{fig:DnD}. The PSNR and SSIM values are also taken from the website. From Figure~\ref{fig:DnD}, it is clear that the methods of~\cite{guo2018CBDnet,zhang2018ffdnet,zhang2017DnCNN} perform poorly in removing the noise from the star and in some cases, the image is over-smoothed. On the other hand, our algorithm can eliminate the noise while preserving the finer details and structures in the star image.

\begin{figure}
\begin{center}
\begin{tabular}[b]{c@{ } c@{ }  c@{ } c@{ } c@{ } c@{ }	}
    \multirow{2}{*}{\includegraphics[width=.15\textwidth,valign=t]{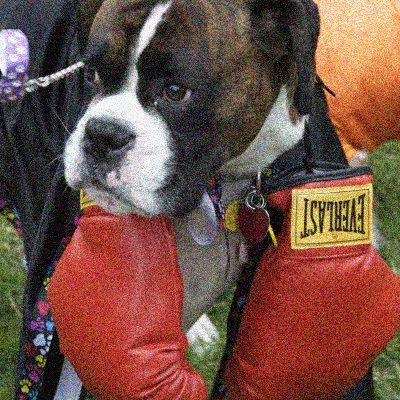}} &  
    \includegraphics[trim={5cm 6cm  1cm  1cm },clip,width=.10\textwidth,valign=t]{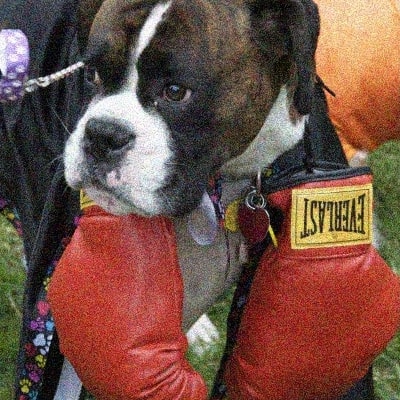}&
  	\includegraphics[trim={5cm 6cm  1cm  1cm },clip,width=.10\textwidth,valign=t]{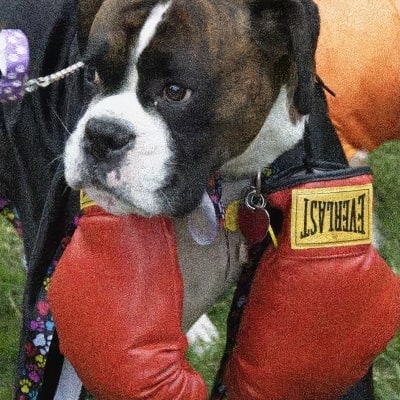}&   
    \includegraphics[trim={5cm 6cm  1cm  1cm },clip,width=.10\textwidth,valign=t]{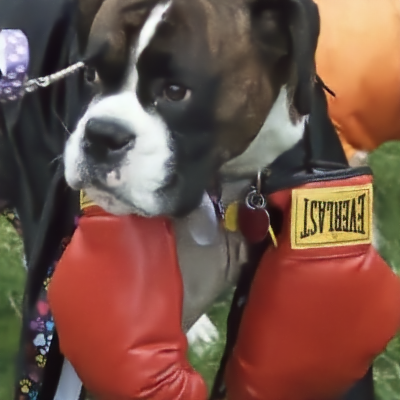}\\
    &
    \includegraphics[trim={1cm 2.2cm  7cm  8cm },clip,width=.10\textwidth,valign=t]{images/RNI15/RNI15_Dog_DnCNN}&
  	\includegraphics[trim={1cm 2.2cm  7cm  8cm 
  	},clip,width=.10\textwidth,valign=t]{images/RNI15/RNI15_Dog_FFD}&   
    \includegraphics[trim={1cm 2.2cm  7cm  8cm },clip,width=.102\textwidth,valign=t]{images/RNI15/RNI15_Dog_Our}\\
    Noisy    & DnCNN     & FFDNet       &  R$^2$Net \\

\end{tabular}
\end{center}
\caption{A real high noise example from RNI15 dataset~\cite{lebrun2015NC}. Our method is able to remove the noise in textured and smooth areas without introducing artifacts.}
\label{fig:NC_high}
\end{figure}

\begin{figure}
\begin{center}
\begin{tabular}[b]{c@{}c@{}c@{}c} 
      
\includegraphics[width=.12\textwidth]{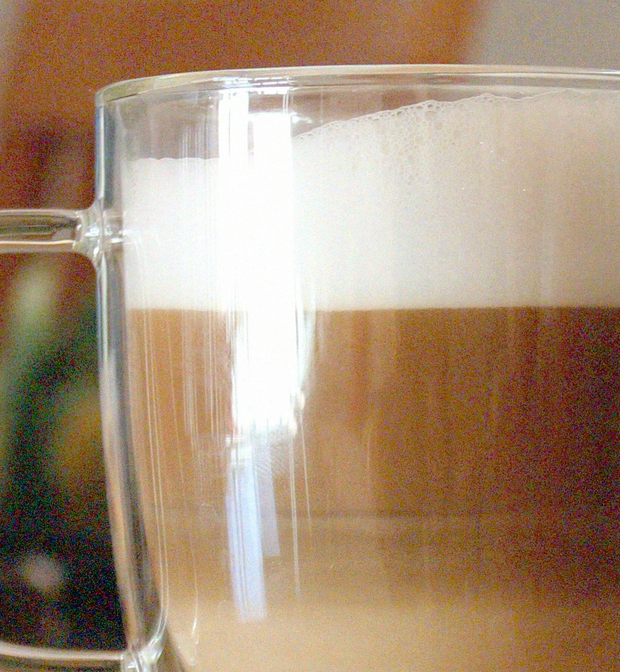}&
\includegraphics[width=.12\textwidth]{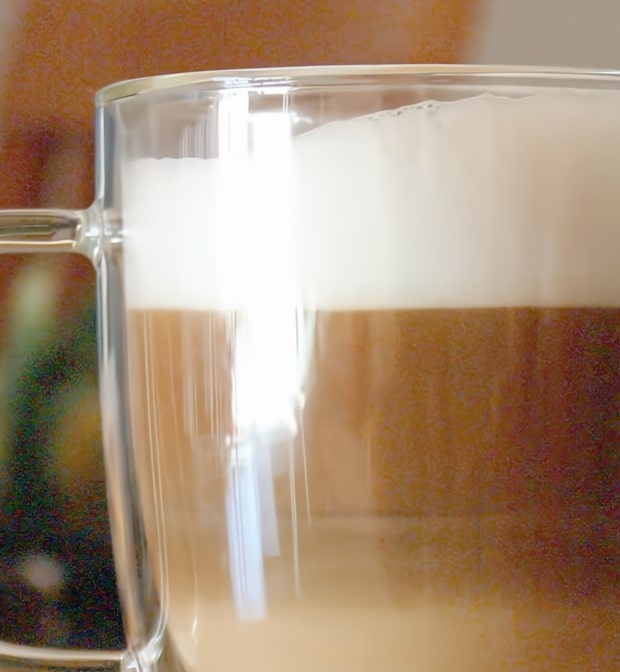}&   
\includegraphics[width=.12\textwidth]{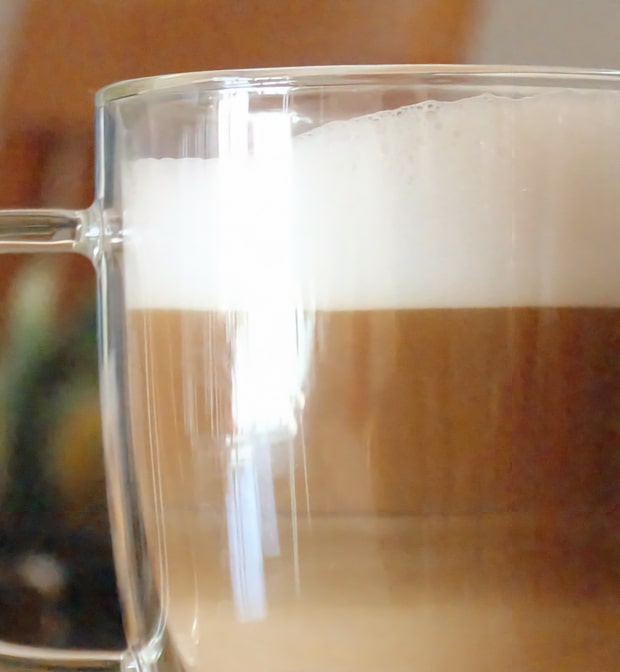}&
\includegraphics[width=.12\textwidth]{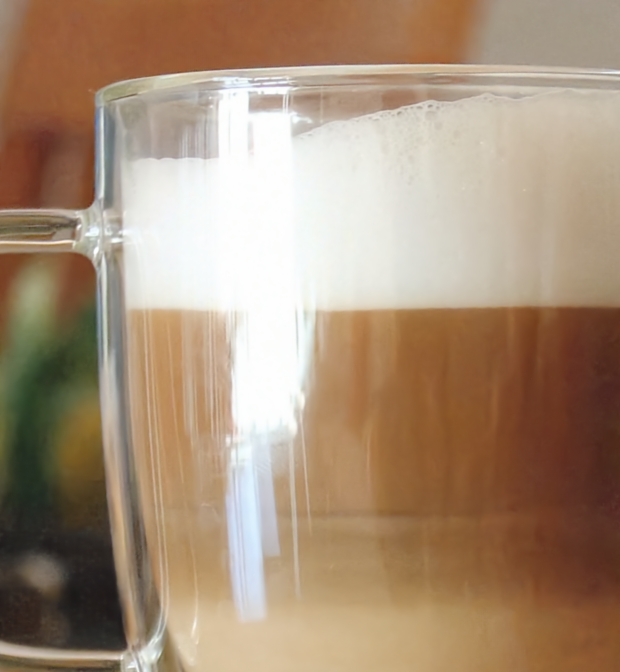}\\
Noisy & FFDNet & CBDNet & R$^2$Net\\
\end{tabular}
\end{center}
\caption{Comparison of our method against the other methods on a real image from RNI15~\cite{lebrun2015NC} benchmark containing spatially variant noise. }
\label{fig:NC}
\end{figure}

\vspace{3mm}
\noindent
\textbf{RNI15}: On RNI15~\cite{lebrun2015NC}, we provide qualitative images only as the ground-truth images are not available. Figure~\ref{fig:NC} presents the denoising results on a low noise intensity image. FFDNet~\cite{zhang2018ffdnet} and CBDNet~\cite{guo2018CBDnet} are unable to remove the noise in its totality, as can been seen near the bottom left of the handle and body of the cup image. On the contrary, our method is able to remove the noise without the introduction of any artifacts. We present another example from the RNI15 dataset~\cite{lebrun2015NC} with high noise in Figure~\ref{fig:NC_high}. CDnCNN~\cite{zhang2017DnCNN} and FFDNet~\cite{zhang2018ffdnet} produce results of limited nature as some noisy elements can be seen in the near the eye and gloves of the Dog image. In comparison, our algorithm recovers the actual texture and structures without compromising on the removal of noise from the images.

\begin{figure}[t]
\begin{center}
\begin{tabular}[b]{c@{ }c@{ }c} 
      
\includegraphics[width=.15\textwidth]{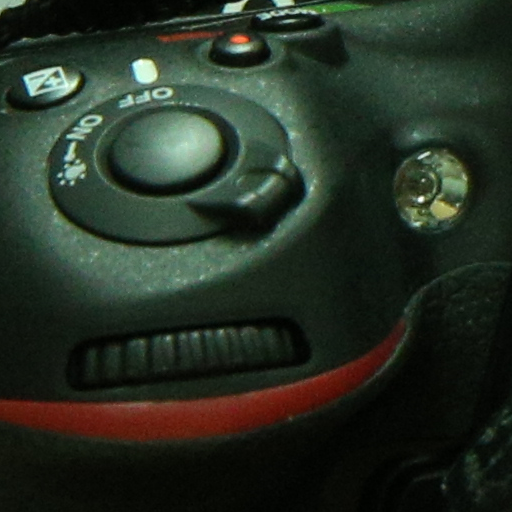}&   
\includegraphics[width=.15\textwidth]{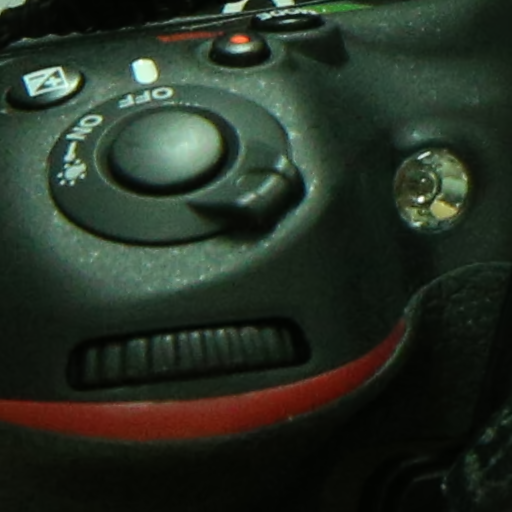}&
\includegraphics[width=.15\textwidth]{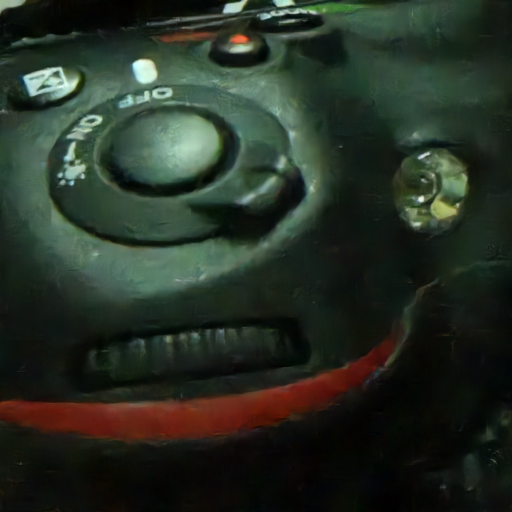}\\
Noisy & CBM3D (39.13) & IRCNN (33.73)\\
\includegraphics[width=.15\textwidth]{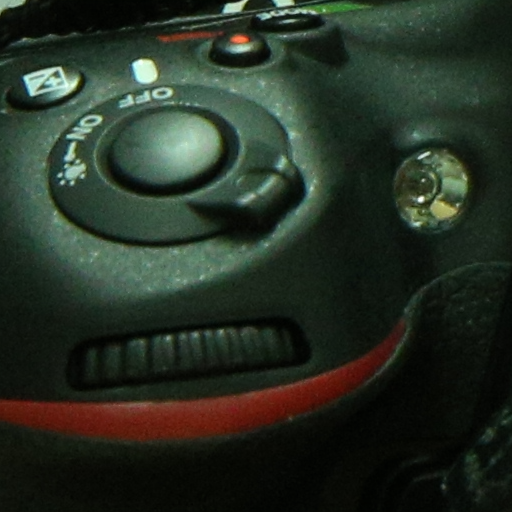}&
\includegraphics[width=.15\textwidth]{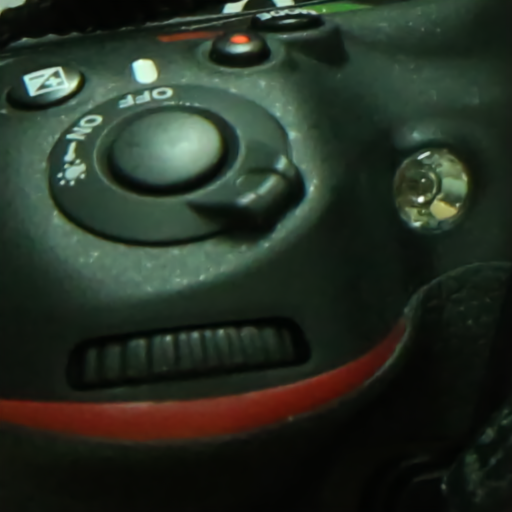}&
\includegraphics[width=.15\textwidth]{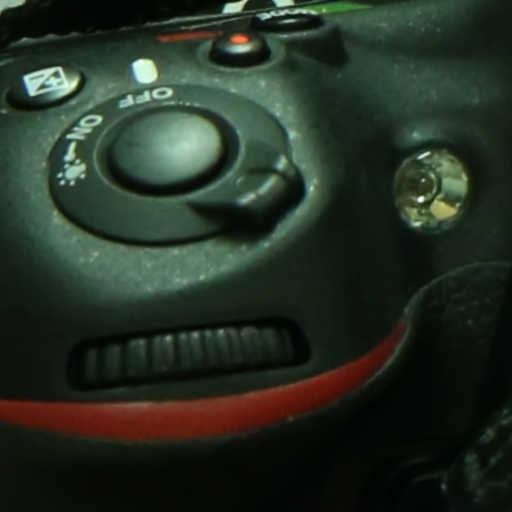}\\
DnCNN (37.56) & CBDNet (40.40) & R$^2$Net (40.50) \\
\end{tabular}
\end{center}
\caption{An image from Nam dataset~\cite{nam2016holistic} with JPEG compression. CBDNet is trained explicitly on JPEG compressed images; still, our performed better.}
\label{fig:Nam}
\end{figure}
\vspace{3mm}
\noindent
\textbf{Nam}: We present the average PSNR scores of the resultant denoised images in Table~\ref{table:Nam_SSID}. Unlike CBDNet~\cite{guo2018CBDnet}, which is trained on Nam~\cite{nam2016holistic} to specifically deal with the JPEG compression, we use the same network to denoise the Nam images~\cite{nam2016holistic} and achieve favorable PSNR numbers. Our performance in terms of PSNR is higher than any of the current state-of-the-art algorithms. Furthermore, our claim is supported by the visual quality of the images produced by our model, as shown in Figure~\ref{fig:Nam}. The amount of noise present after denoising by our method is negligible as compared to CDnCNN and other counterparts.  

\begin{figure}
\begin{center}
\begin{tabular}[b]{c@{ }c@{ }c@{ }c} 
      
\includegraphics[width=.11\textwidth]{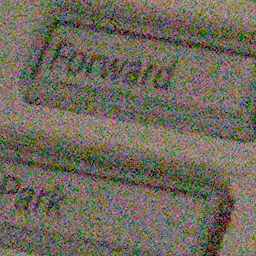}&   
\includegraphics[width=.11\textwidth]{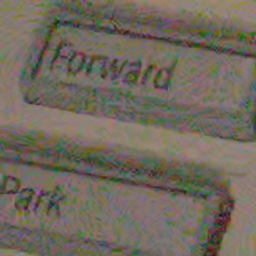}&
\includegraphics[width=.11\textwidth]{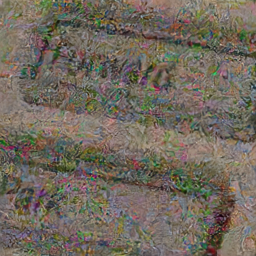}&
\includegraphics[width=.11\textwidth]{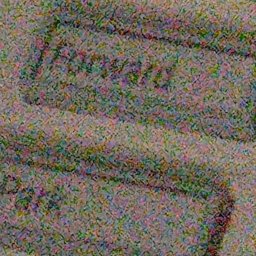}\\
 & 25.75 dB& 21.97 dB& 20.76 dB\\
Noisy & CBM3D & IRCNN  & DnCNN \\

\includegraphics[width=.11\textwidth]{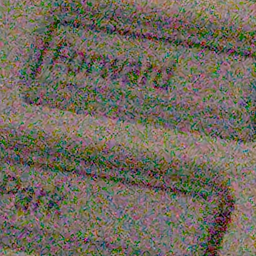}&
\includegraphics[width=.11\textwidth]{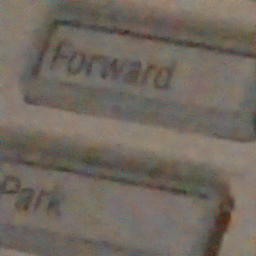}&
\includegraphics[width=.11\textwidth]{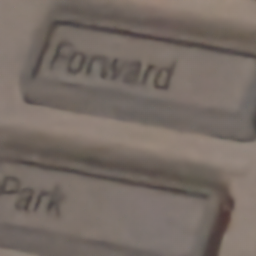}&
\includegraphics[width=.11\textwidth]{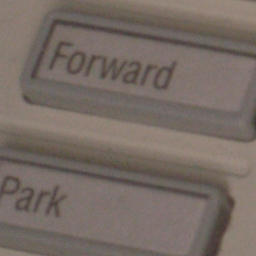}\\
19.70 dB & 28.84 dB& \textbf{35.57 dB}&  \\
FFDNet  & CBDNet  & Ours& GT \\

\end{tabular}
\end{center}
\caption{A challenging example from SSID dataset~\cite{abdelhamed2018high}. Our method can remove noise and restore true colors.}
\label{fig:SSID}
\end{figure}

\vspace{3mm}
\noindent
\textbf{SSID}: As the last dataset, we employ the SSID real noise dataset, which has the highest number of test (validation) images available. The results in terms of PSNR are shown in the second row of Table~\ref{table:Nam_SSID}. Again, it is clear that our method outperforms FFDNet~\cite{zhang2018ffdnet} and CBDNet~\cite{guo2018CBDnet} by a margin of 9.5dB and 7.93dB, respectively.  In Figure~\ref{fig:SSID}, we show the denoised results of a challenging image by different algorithms. Our technique recovers the true colors closer to the original pixel values while competing methods are unable to restore original colors and, in specific regions, induce false colors.

\begin{table*}[t]
\caption{The performance of super-resolution algorithms on  Set5, Set14, BSD100, and URBAN100 datasets for upscaling factors of 2, 3, and 4. The bold highlighted results are the best on single image super-resolution. }
\centering
\resizebox{\textwidth}{!}{%
\begin{tabular}{l|c|c|c|c|c|c|c|c|c}
\hline\hline
Dataset	 	& Scale             &Bicubic		&TNRD	        & VDSR		    &DnCNN 			&SRMD 				&CARN            &R$^2$Net           & R$^2$Net+\\ \hline \hline
			&	$\times$2		&33.66 / 0.9299	&36.86 / 0.9556	&37.56 / 0.9591	&37.58 / 0.9590 &37.79 / 0.9601     &37.76 / 0.9590  &37.95 / 0.9605 &\textbf{38.07} / \textbf{0.9608}\\
Set5	 	&   $\times$3		&30.39 / 0.8682	&33.18 / 0.9152	&33.67 / 0.9220	&33.75 / 0.9222	&34.12 / 0.9254     &34.29 / 0.9255  &34.37 / 0.9269 &\textbf{34.49} / \textbf{0.9278}\\
			&	$\times$4		&28.42 / 0.8104	&30.85 / 0.8732	&31.35 / 0.8845	&31.40 / 0.8845	&31.96 / 0.8925     &32.13 / 0.8937  &32.15 / 0.8946 &\textbf{32.34} / \textbf{0.8970}\\ \hline \hline
			&	$\times$2		&30.24 / 0.8688	&32.51 / 0.9069	&33.02 / 0.9128	&33.03 / 0.9128	&33.32 / 0.9159		&33.52 / 0.9166  &33.54 / 0.9173 &\textbf{33.63} / \textbf{0.918}2\\
Set14		&	$\times$3		&27.55 / 0.7742	&29.43 / 0.8232	&29.77 / 0.8318	&29.81 / 0.8321	&30.04 / 0.8382		&30.29 / 0.8407  &30.34 / 0.8419 &\textbf{30.43} / \textbf{0.8433}\\
			&	$\times$4		&26.00 / 0.7027	&27.66 / 0.7563	&27.99 / 0.7659	&28.04 / 0.7672	&28.35 / 0.7787		&28.60 / 0.7806  &28.62 / 0.7822 &\textbf{28.72} / \textbf{0.7842}\\ \hline \hline
			&	$\times$2		&29.56 / 0.8431	&31.40 / 0.8878	&31.89 / 0.8961	&31.90 / 0.8961	&32.05 / 0.8985		&32.09 / 0.8978  &32.19 / 0.9001 &\textbf{32.25} / \textbf{0.9007}\\
BSD100		&	$\times$3		&27.21 / 0.7385	&28.50 / 0.7881	&28.82 / 0.7980	&28.85 / 0.7981	&28.97 / 0.8025		&29.06 / 0.8034  &29.12 / 0.8055 &\textbf{29.17} / \textbf{0.8065}\\
			&	$\times$4		&25.96 / 0.6675	&27.00 / 0.7140	&27.28 / 0.7256	&27.29 / 0.7253	&27.49 / 0.7337		&27.58 / 0.7349  &27.60 / 0.7363 &\textbf{27.6}5 / \textbf{0.7376}\\ \hline \hline
			&	$\times$2		&26.88 / 0.8403	&29.70 / 0.8994	&30.76 / 0.9143	&30.74 / 0.9139	&31.33 / 0.9204		&31.92 / 0.9256  &32.07 / 0.9280 &\textbf{32.24} / \textbf{0.9294}\\
Urban100	&	$\times$3		&24.46 / 0.7349	&26.42 / 0.8076	&27.13 / 0.8283	&27.15 / 0.8276	&27.57 / 0.8398		&28.06 / 0.8493  &28.14 / 0.8519 &\textbf{28.28} / \textbf{0.8542}\\
			&	$\times$4		&23.14 / 0.6577	&24.61 / 0.7291	&25.17 / 0.7528	&25.20 / 0.7521	&25.68 / 0.7731		&26.07 / 0.7837  &26.18 / 0.7881 &\textbf{26.28} / \textbf{0.7905}\\\hline

\end{tabular}
}
\label{table:SISR}
\end{table*}

\subsection{Super-resolution Comparisons}

\begin{figure*}[t]
\begin{center}
\begin{tabular}[b]{c@{ } c@{ }  c@{ } c@{ } c@{ } c@{ }	c}
    \multirow{4}{*}{\includegraphics[trim={3cm 16cm  3cm  0cm },clip,width=.24\textwidth,valign=t]{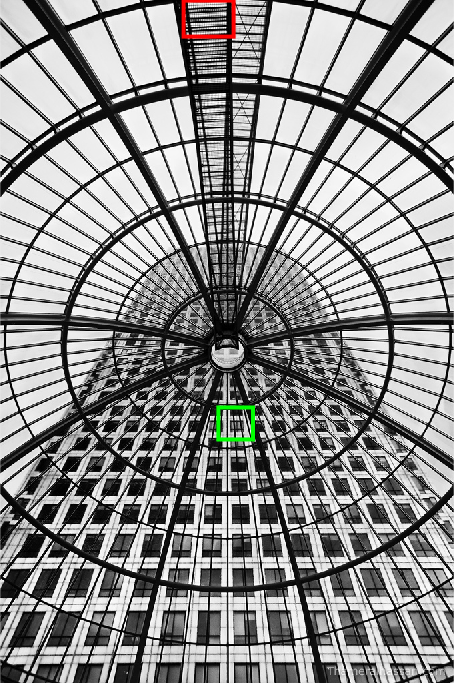}} &  
    \includegraphics[width=.12\textwidth,valign=t]{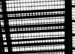}&
  	\includegraphics[width=.12\textwidth,valign=t]{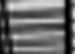}&
  	\includegraphics[width=.12\textwidth,valign=t]{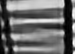}&
    \includegraphics[width=.12\textwidth,valign=t]{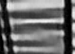}&
    \includegraphics[width=.12\textwidth,valign=t]{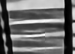}&
	\includegraphics[width=.12\textwidth,valign=t]{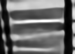}\\
    & GT & Bicubic & SRCNN& FSRCNN & DRRN & DRCN\\

    &
    \includegraphics[width=.12\textwidth,valign=t]{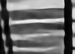}&
  	\includegraphics[width=.12\textwidth,valign=t]{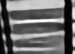}&   
    \includegraphics[width=.12\textwidth,valign=t]{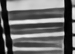}&
    \includegraphics[width=.12\textwidth,valign=t]{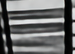}&
	\includegraphics[width=.12\textwidth,valign=t]{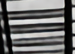}&
    \includegraphics[width=.12\textwidth,valign=t]{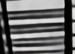}\\
  Urban100 \enquote{img\_72} & VDSR & LapSRN & MS-LapSRN & CARN &R$^2$Net & R$^2$Net+\\
\end{tabular}
\end{center}
\caption{The visual comparisons for 4$\times$ super-resolution against several state-of-the-art algorithms on an image from Urban100~\cite{huang2015URBAN100} dataset. Our R$^2$Net results are the most accurate.}
\label{fig:SISR}
\vspace*{-3mm}
\end{figure*}

We evaluate our model concerning networks that aim for efficiency as well as PSNR numbers and having similar depth and number of parameters. We compare against TNRD~\cite{chen2017TNRD}, SRMD~\cite{zhang2018SRMDNF}, CARN~\cite{ahn2018CARN}\footnote{CARN has more than hundred convolutional layers} \etc, as opposed to RCAN~\cite{zhang2018RCAN}, DRLN~\cite{anwar2019densely} which have more than 16M parameters while our model has only 1.49M parameters.  We present the performance on four publicly available datasets given below:
\begin{itemize}
\item Set5~\cite{bevilacqua2012Set5} is a classical dataset; contains only five images.
\item Set14~\cite{zeyde2010Set14} contains 14 RGB images.
\item BSD100~\cite{Martin2001BSD} is the subset of the Berkely Segmentation dataset and consists of one hundred natural images.
\item Urban100~\cite{huang2015URBAN100} is a recently proposed dataset of 100 images. The images contain human-made objects and buildings. The size of the image and the structures present in the dataset make it very challenging for the super-resolution task.
\end{itemize}
\begin{table}[t]
\caption{Quantitative results for the SSID~\cite{abdelhamed2018high} \& Nam~\cite{nam2016holistic}.}
\centering
\resizebox{\columnwidth}{!}{%
\begin{tabular}{l|c|c|c|c|c}
\hline\hline
  \multicolumn{6}{c}{Methods} \\ \hline
Datasets    &BM3D     &  DnCNN   & FFDNet	 & CBDNet    & Ours	  \\\hline 
Nam~\cite{nam2016holistic}         &37.30    &  35.55   & 38.7    & 39.01     &\textbf{39.09} \\ 
SSID~\cite{abdelhamed2018high}        &30.88    &  26.21   & 29.20   & 30.78     &\textbf{38.71} \\ \hline \hline
\end{tabular}}
\label{table:Nam_SSID}
\end{table}

\subsubsection{R$^2$Net for SR} 
\begin{figure}[t]
\begin{center}
\includegraphics[clip, trim=9.5cm 10cm 8cm 0cm, width=\columnwidth]{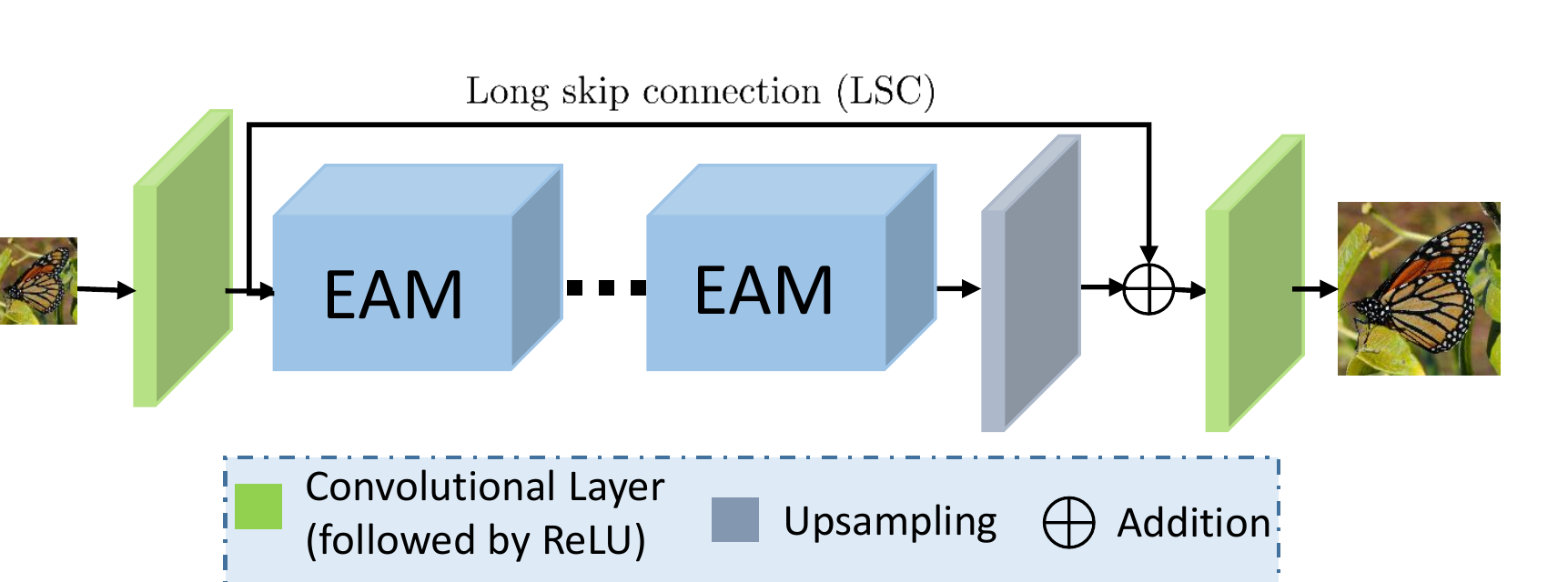}\\ 
\end{center}
\caption {The architecture for super-resolution with two modifications: the change in position of long skip connection and insertion of upsampling layer.}
\label{fig:SRR2Net}
\vspace*{-3mm}
\end{figure}

The R$^2$Net is modified for super-resolution due to the increase in the size of the final output. There are two modifications performed in the network structure: 1) An additional layer (upsampling layer) is inserted before the final convolutional layer to super-resolve the input image to the desired resolution, 2) the residual learning is performed by changing the position of the long skip connection ( \ie output of the feature extraction layer is added to the output of final EAM block), as shown in Figure~\ref{fig:SRR2Net}. The second modification is due to the change in the size of the features after upsampling. It is also necessary to mention that the network's input size is 48$\times$48 for super-resolution. Except for the mentioned modifications, no additional changes are made to the network. 

\subsubsection{Visual Comparisons} 
We furnish an image from Urban100~\cite{huang2015URBAN100} for qualitative comparison in Figure~\ref{fig:SISR} against the algorithms which have a similar number of parameters and aim to provide efficient solutions for super-resolution. It is evident from Figure that all the competing methods fail to recover the straight lines forming rectangles shown in the cropped regions from \enquote{img\_72} image. Our method performs better than current state-of-the-art CARN~\cite{ahn2018CARN}. Moreover, our network produces results that are faithful to the ground-truth image and without any blurring.

\subsubsection{Quantitative Comparisons}
Table~\ref{table:SISR} shows the average PSNR and SSIM values for the mentioned four datasets against several state-of-the-art algorithms. Our algorithm outperforms all the methods for all scaling factors on all datasets. Our method is very lightweight compared to recently published CARN~\cite{ahn2018CARN} \ie contains 4$\times$ fewer layers compared to CARN~\cite{ahn2018CARN}. Our network's performance becomes comparatively better when the number of images in testing datasets and the scaling factor increases. Similarly, our R$^2$Net achieves significantly better results than the VDSR~\cite{kim2016VDSR}, which was state-of-the-art until a year ago.

\subsection{Raindrop Removal Comparisons}
In this section, We present the performance of R$^2$Net on raindrop removal and compare with three state-of-the-art algorithms which include Eigen~\etal~\cite{eigen2013restoring}, Pix2Pix~\cite{isola2017image} and DeRain~\cite{qian2018DeRain} on two test datasets introduced in~\cite{qian2018DeRain} termed as \enquote{Test\_a} and \enquote{Test\_b}.  We use the same number of training images as DeRain~\cite{qian2018DeRain}; however, we train using cropped patches instead of whole images. 

\begin{table}[t]
\caption{The average PSNR(dB)/SSIM from different methods on raindrop~\cite{qian2018DeRain} dataset.}
\centering
\resizebox{\columnwidth}{!}{%
\begin{tabular}{l|c|c|c|c|}
\hline\hline
Datasets    	& Eigen			 & Pix2pix		    & DeRain         & R$^2$Net (Ours) \\ \hline \hline
Test\_a			& 28.59 / 0.6726 & 30.14 / 0.8299	& 31.57 / 0.9023 & \textbf{32.03} / \textbf{0.9325}\\
Test\_b			&	-			 &	-			    & 24.93 / 0.8091 & \textbf{26.42} / \textbf{0.8255}\\ \hline
\end{tabular}}
\label{table:DeRain}
\end{table}

\begin{figure}[t]
\begin{center}
\begin{tabular}[b]{c@{ } c@{ }  c@{ }}
    \multirow{4}{*}{\includegraphics[trim={0cm 0.7cm  4.3cm  0cm },clip,width=.224\textwidth,valign=t]{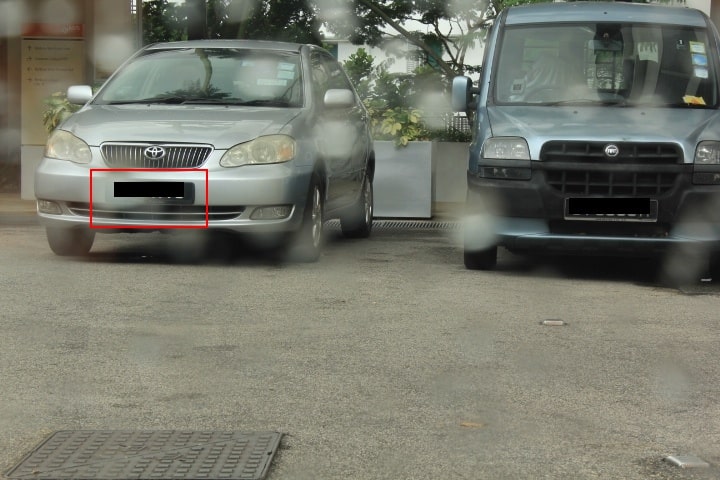}} &  
    \includegraphics[width=.125\textwidth,valign=t]{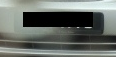}&
	\includegraphics[width=.125\textwidth,valign=t]{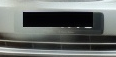}\\
    & 22.25dB  & 28.14dB   \\
    & Rain      & DeRain~\cite{qian2018DeRain} \\

    &
     \includegraphics[width=.125\textwidth,valign=t]{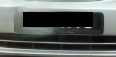}&   
     \includegraphics[width=.125\textwidth,valign=t]{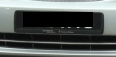}\\

                  & 29.14dB    &   \\
     Rain Image  & R$^2$Net    &  GT \\
     \end{tabular}
\end{center}
\vspace{-2mm}
\caption{The visual comparisons on a rainy image. The figure is showing the plate which is affected by raindrops. Our method is consistent in restoring raindrop affected areas.}
\label{fig:DeRain1}

\begin{center}
\begin{tabular}[b]{c@{ } c@{ }  c@{ }}
         \multirow{4}{*}{\includegraphics[trim={0cm 0cm  12cm  0cm },clip,width=.224\textwidth,valign=t]{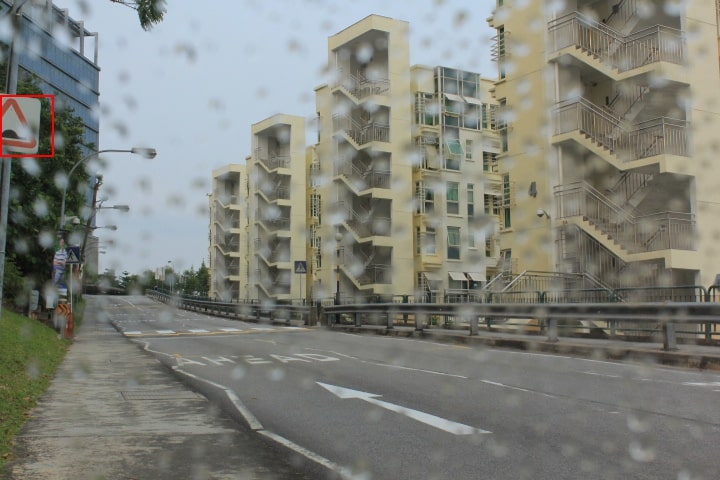}} &  
    \includegraphics[trim={0cm 0.2cm  0cm  0.2cm },clip,width=.125\textwidth,valign=t]{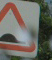}&
	\includegraphics[trim={0cm 0.2cm  0cm  0.2cm },clip,width=.125\textwidth,valign=t]{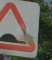}\\
    & 24.73dB  &  29.35dB   \\
    & Rain      & DeRain~\cite{qian2018DeRain} \\

    &
     \includegraphics[trim={0cm 0.2cm  0cm  0.20cm },clip,width=.125\textwidth,valign=t]{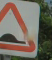}&   
     \includegraphics[trim={0cm 0.2cm  0cm  0.2cm },clip,width=.125\textwidth,valign=t]{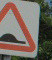}\\

                  & 30.93dB    &     \\
     Rain Image  & R$^2$Net    &  GT \\
    
\end{tabular}
\end{center}
\vspace*{-2mm}
\caption{Another example of a rainy image. The cropped region is showing the road sign affected by raindrops. Our method recovers the distorted colors closer to the ground-truth.}
\label{fig:DeRain2}
\end{figure}

\subsubsection{Visual Comparisons}
Figure~\ref{fig:DeRain1} present an example image from the \enquote{Test\_b} dataset, showing the cropped region near the front end of the car. DeRain~\cite{qian2018DeRain} fails to remove the effect of the raindrop and results in the same image as the input. On the other hand, our method restores the edges in the input rainy image. 

The second example in Figure~\ref{fig:DeRain2} shows a rainy urban scene. We focus and crop the road sign to visualize better the differences between the output of R$^2$Net and competing methods. The DeRain~\cite{qian2018DeRain} network removes the edges and color information where the raindrop affected the road sign. In our case, the edges and the color both are restored and are closer to the ground-truth clean image.  

\subsubsection{Quantitative Comparisons}
Table~\ref{table:DeRain} presents the quantitative results on both \enquote{Test\_a} and \enquote{Test\_b} for the mentioned algorithms. Compared to recent state-of-the-art in rain drop removal \ie DeRain~\cite{qian2018DeRain}, our gain is 0.46dB for \enquote{Test\_a} and a significant improvement of 1.49dB on the challenging \enquote{Test\_b}. Similarly, the gain from Eigen~\etal~\cite{eigen2013restoring} is about 3.44dB. These results illustrate that our method can restore images that are similar in structure to the corresponding ground-truth images.

\subsection{JPEG Comparisons}
For JPEG compression, our method is compared against three competing methods, which include AR-CNN~\cite{dong2015ARCNN}, TNRD~\cite{chen2017TNRD}, and DnCNN~\cite{zhang2017DnCNN}. All models are trained for four quality factors of 10, 20, 30, and 40 except for TNRD~\cite{chen2017TNRD}, which is only trained for the first three JPEG quality factors. 

\begin{table*}[t]
\caption{Average PSNR/SSIM for JPEG image deblocking for quality factors of 10, 20, 30, and 40 on LIVE1~\cite{sheikh2005live1} dataset. The best results are in bold.}
\centering
\begin{tabular}{l|c|c|c|c|c|c|c}
\hline\hline
Quality  &LD~\cite{li2014LD}  			&AR-CNN~\cite{dong2015ARCNN}        &TNRD~\cite{chen2017TNRD}	        &DnCNN~\cite{zhang2017DnCNN}                    &LPIO~\cite{fan2018LPIO}        &DCSC~\cite{fu2019DCSC}		      & R$^2$Net\\ \hline\hline				
		
10	     &28.26 / 0.805 &28.98 / 0.8076	&29.15 / 0.8111	&\textbf{29.19} / 0.8123   &29.17/0.811 &29.17/0.815  & \textbf{29.17} / \textbf{0.8202}\\
20	     &30.19 / 0.871 &31.29 / 0.8733	&31.46 / 0.8769	&31.59 / 0.8802 		   &31.52/0.876 &31.48/0.880  & \textbf{32.28} / \textbf{0.8957}\\
30	     &31.32 / 0.898 &32.67 / 0.9043	&32.84 / 0.9059	&32.98 / 0.9090 		   &32.99/0.907 &32.83/0.909  & \textbf{33.61} / \textbf{0.9206}\\
40	     &-             &33.63 / 0.9198	&-				&33.96 / 0.9247 		   & -		    &  -          & \textbf{34.66} / \textbf{0.9352}\\ \hline
\end{tabular}
\label{table:JPEG}
\end{table*}	

\begin{figure}
\begin{center}
\begin{tabular}[b]{c@{ } c@{ }  c@{ }}
    \multirow{4}{*}{\includegraphics[trim={8cm 0.9cm  4.3cm  0.4cm },clip,width=.224\textwidth,valign=t]{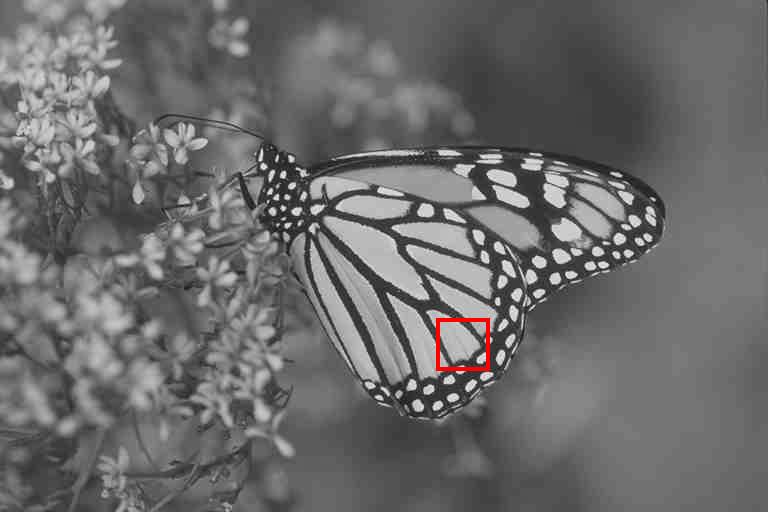}} &  
    \includegraphics[trim={0cm 0cm  0cm  0.3cm },clip,width=.125\textwidth,valign=t]{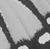}&
	\includegraphics[trim={0cm 0cm  0cm  0.23cm },clip,width=.125\textwidth,valign=t]{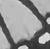}\\
    &         & 37.22dB   \\
    & GT      & ARCNN~\cite{dong2015ARCNN} \\

    &
     \includegraphics[trim={0cm 0cm  0cm  0.3cm },clip,width=.125\textwidth,valign=t]{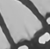}&   
     \includegraphics[trim={0cm 0cm  0cm  0.3cm },clip,width=.125\textwidth,valign=t]{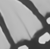}\\

                  & 37.20dB    & 39.27dB  \\
     JPEG Monarch Image  & DnCNN~\cite{zhang2017DnCNN}    &  R$^2$Net \\
    
\end{tabular}
\end{center}
\vspace*{-2mm}
\caption{A sample image of a Monarch with the artifacts having a quality factor of 20. Our R$^2$Net restore texture correctly, specifically the line, as shown in the zoomed version of the restored patch.}
\label{fig:JPEG1}
\end{figure}

\subsubsection{Visual Comparisons}
In Figure~\ref{fig:JPEG1}, we show a comparison of our method on the \enquote{Monarch} image. Our network can retrieve the fine details such as the straight line in the wing shown in the close up while ARCNN~\cite{dong2015ARCNN} and DnCNN~\cite{zhang2017DnCNN} fail to achieve the desired results and produce distorted lines. 

\begin{figure}
\begin{center}
\begin{tabular}[b]{c@{ } c@{ }  c@{ }}
    \multirow{4}{*}{\includegraphics[trim={0cm 0.2cm  10cm  0.2cm },clip,width=.224\textwidth,valign=t]{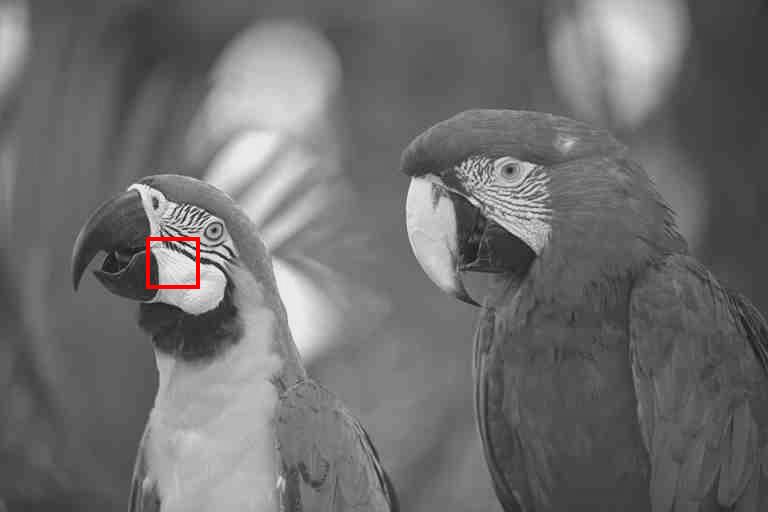}} &  
    \includegraphics[trim={0cm 0cm  0cm  0.4cm },clip,width=.125\textwidth,valign=t]{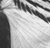}&
	\includegraphics[trim={0cm 0cm  0cm  0.3cm },clip,width=.125\textwidth,valign=t]{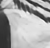}\\
    &     & 34.34dB   \\
    & GT      & ARCNN~\cite{dong2015ARCNN} \\

    &
     \includegraphics[trim={0cm 0cm  0cm  0.4cm },clip,width=.125\textwidth,valign=t]{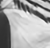}&   
     \includegraphics[trim={0cm 0cm  0cm  0.4cm },clip,width=.125\textwidth,valign=t]{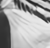}\\

&34.20dB    & 35.73dB  \\
     JPEG Parrot Image  & DnCNN~\cite{zhang2017DnCNN}    &  R$^2$Net \\
    
\end{tabular}
\end{center}
\vspace*{-2mm}
\caption{A different example of the artifact image removal for a quality factor of 20. R$^2$Net restores the texture accurately on the face of the parrot.}
\label{fig:JPEG2}
\vspace*{-5mm}
\end{figure}

Similarly, in  Figure~\ref{fig:JPEG2}, the \enquote{Parrot} image, it can be observed that our model output has fewer artifacts and restores structures more accurately on the face of the parrot. On the other hand, ARCNN~\cite{dong2015ARCNN} and DnCNN~\cite{zhang2017DnCNN} smooth out the texture and lines present in the ground-truth images. These outcomes show the importance of our attention mechanism and the enhanced capacity of the proposed model.

\subsubsection{Quantitative Comparisons}
The JPEG deblocking average results in terms of PSNR and SSIM are listed in Table~\ref{table:JPEG} for different methods. Our gain over ARCNN~\cite{dong2015ARCNN} and DnCNN~\cite{zhang2017DnCNN} for a compression factor of 40 is significant \ie 1.03dB and  0.7dB, respectively. Similarly, the overall improvement on the LIVE1 dataset for R$^2$Net is 0.79dB (over~\cite{dong2015ARCNN}) and 0.5dB (over~\cite{zhang2017DnCNN}) for all compression factors.

 Moreover, DCSC~\cite{fu2019DCSC} is specifically designed to remove JPEG artifacts and propose it in parallel with our preliminary version (RIDNet).  As shown in Table~\ref{table:JPEG}, we have outperformed DCSC by order of magnitude in terms of PSNR and SSIM. This shows the modeling power of our network for image restoration.

\section{Conclusion}
In this paper, we present a new CNN restoration model for real degraded photographs. This is the first end-to-end single pass network to show state-of-the-art results across a broad range of real image restoration and enhancement tasks. Specifically, we show results on denoising, super-resolution, raindrop removal, and compression artifacts.

Unlike previous algorithms, our model is a single-blind restoration network for real degraded images. We propose a new restoration module to further learn the features and enhance the network's capability further; we adopt feature attention to rescale the channel-wise features by taking into account the dependencies between the channels. We also use LSC, SSC, and SC to allow low-frequency information to bypass so the network can focus on residual learning.  Extensive experiments on 11 real-degraded datasets for four restoration tasks against more than 30 state-of-the-art algorithms demonstrate our proposed model's effectiveness.
\bibliographystyle{spmpsci}      
\bibliography{ref.bib}   

\end{document}